\documentclass[sigconf]{acmart}
\usepackage{url}
\usepackage[utf8]{inputenc}
\usepackage[]{caption}
\usepackage{graphicx}
\usepackage{amsmath}
\usepackage{amsthm}
\usepackage{booktabs}
\usepackage{algorithm}
\usepackage{algorithmicx}
\usepackage{algpseudocode}
\usepackage{amsmath}
\usepackage{pifont}

\newcommand{\xmark}{\text{\ding{55}}}
\usepackage{xcolor,pifont}
\newcommand*\colourcheck[1]{%
  \expandafter\newcommand\csname #1check\endcsname{\textcolor{#1}{\ding{52}}}%
}
\usepackage{multirow}
\usepackage{adjustbox}
\usepackage{amsmath}
\usepackage{enumitem}
\DeclareMathOperator*{\argmax}{arg\,max}

\newcommand{\proposed}{\textsf{LTE4G}}
\usepackage{amsmath}

\usepackage{algpseudocode}%
\usepackage{balance} 

\AtBeginDocument{%
  \providecommand\BibTeX{{%
    \normalfont B\kern-0.5em{\scshape i\kern-0.25em b}\kern-0.8em\TeX}}}

\copyrightyear{2022}
\acmYear{2022}
\setcopyright{acmcopyright}
\acmConference[CIKM '22] {Proceedings of the 31st ACM International Conference on Information and Knowledge Management}{October 17--21, 2022}{Atlanta, GA, USA.}
\acmBooktitle{Proceedings of the 31st ACM International Conference on Information and Knowledge Management (CIKM '22), October 17--21, 2022, Atlanta, GA, USA}
\acmPrice{15.00}
\acmISBN{978-1-4503-9236-5/22/10}
\acmDOI{10.1145/3511808.3557381}

\begin{document}

\title{LTE4G: Long-Tail Experts for Graph Neural Networks}

\author{Sukwon Yun}
\affiliation{%
  \institution{KAIST ISysE}
  \city{Daejeon}
  \state{}
  \country{Republic of Korea}
  \postcode{34141}
}
\email{swyun@kaist.ac.kr}

\author{Kibum Kim}
\affiliation{%
  \institution{KAIST ISysE}
  \city{Daejeon}
  \state{}
  \country{Republic of Korea}
  \postcode{34141}
}
\email{rlqja1107@kaist.ac.kr}

\author{Kanghoon Yoon}
\affiliation{%
  \institution{KAIST ISysE}
  \city{Daejeon}
  \state{}
  \country{Republic of Korea}
  \postcode{34141}
}
\email{ykhoon08@kaist.ac.kr}

\author{Chanyoung Park}
\authornote{Corresponding author}
\affiliation{%
  \institution{KAIST ISysE \& AI}
  \city{Daejeon}
  \state{}
  \country{Republic of Korea}
  \postcode{34141}
}
\email{cy.park@kaist.ac.kr}

\renewcommand{\shortauthors}{Sukwon Yun, Kibum Kim, Kanghoon Yoon, \& Chanyoung Park}

\begin{abstract}
\looseness=-1
Existing Graph Neural Networks (GNNs) usually assume a balanced situation where both the class distribution and the node degree distribution are balanced. However, in real-world situations, we often encounter cases where a few classes (i.e., head class) dominate other classes (i.e., tail class) as well as in the node degree perspective, and thus naively applying existing GNNs eventually fall short of generalizing to the tail cases. Although recent studies proposed methods to handle long-tail situations on graphs, they only focus on either the class long-tailedness or the degree long-tailedness. 
In this paper, we propose a novel framework for training GNNs, called \textsf{L}ong-\textsf{T}ail \textsf{E}xperts for \textsf{G}raphs (\proposed), which jointly considers the class long-tailedness, and the degree long-tailedness for node classification. 
The core idea is to assign an expert GNN model to each subset of nodes that are split in a balanced manner considering both the class and degree long-tailedness.
After having trained an expert for each balanced subset, we adopt knowledge distillation to obtain two class-wise students, i.e., Head class student and Tail class student, each of which is responsible for classifying nodes in the head classes and tail classes, respectively. 
We demonstrate that~\proposed~outperforms a wide range of state-of-the-art methods in node classification evaluated on both manual and natural imbalanced graphs.
The source code of~\proposed~can be found at \url{https://github.com/SukwonYun/LTE4G}.


\end{abstract}





\maketitle

\section{Introduction}
Graph Neural Networks (GNNs) have achieved astonishing success thanks to their ability to encode graph-structured data containing nodes (objects) and edges (relationships). Such graph-structured data range from social networks, protein-protein interaction (PPI) networks to citation networks. A representative downstream task for GNN is node classification \cite{gcn, graphsage, graphsmote} whose goal is to correctly classify to which class each node belongs to. 
Specifically, in citation networks, papers are represented as nodes and their citation relationships are represented as edges, and the ultimate goal is to accurately predict the category of each paper.

\begin{figure}[!t]
\includegraphics[width=0.9\linewidth]{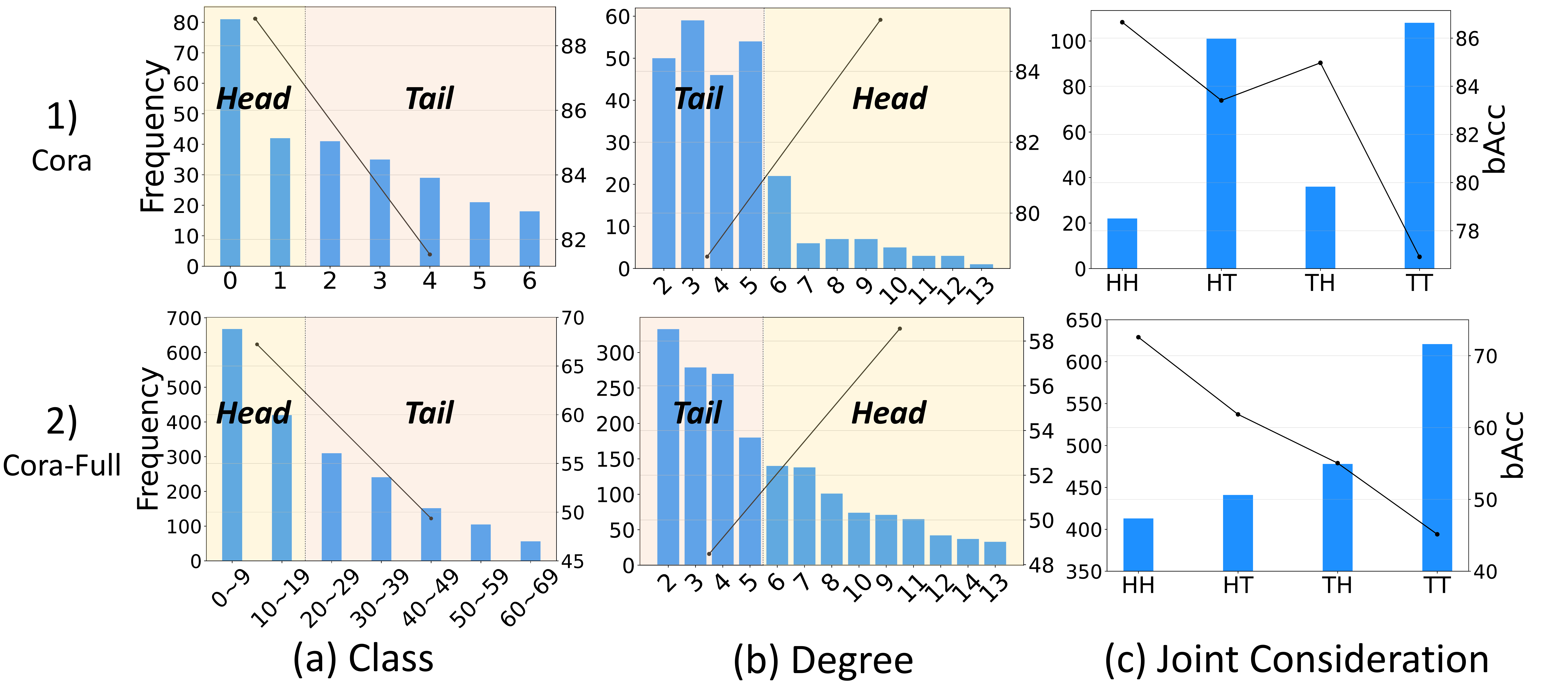}

\vspace{-2ex}
\caption{(a) Class Frequency (b) Degree Frequency, and (c) Frequency of their joint subsets, all reported with balanced accuracy (bAcc) (\%) for node classification.}
\label{fig:figure1}
\vspace{-4ex}
\end{figure}

\looseness=-1
Despite the recent renaissance of GNNs, most existing GNNs assume a balanced situation where the number of nodes per class, and the number of neighbors for each node (i.e., node degree) are balanced. However, as most real-world graphs follow long-tailed distribution in the perspective of both class and node degree, these two key challenges should not be overlooked.
First, the \textit{class long-tailed situation} in which the number of nodes in a few classes dominates that in other classes incurs the GNNs to be biased towards classes with abundant samples (i.e., head class), and thus the GNNs do not generalize well on classes with few samples (i.e., tail class)~\cite{graphsmote,graphens}. 
For example, in the extended Cora citation network (i.e., Cora-Full~\cite{cora-full}) that contains 70 classes, the first ten classes sorted by the number of nodes in each class have 11 times (i.e., $672/60 = 11.2$) more nodes for training than the last ten classes and such tendency can be found in Cora~\cite{citation} dataset as well. Such an imbalance situation negatively affects the performance of GNNs, as depicted in Figure~\ref{fig:figure1}(a). What is worse is that when the number of samples in a tail class is scarce but significant such as in fraud detection \cite{fraud1, fraud2} and bot detection \cite{bot}, GNNs trained on a dataset with class long-tailed distribution are likely to fail in detecting such significant tail cases.

Second, the \textit{degree long-tailed situation} in which the node degree ranges from very high (i.e., head degree) to very low cases (i.e., tail degree) 
incurs the GNNs to be biased towards nodes with high degree~\cite{tailgnn,metatail2vec} (Figure~\ref{fig:figure1}(b)). 
More precisely, due to the neighborhood aggregation scheme of GNNs, nodes that are linked with many other nodes tend to have higher-quality representations, while nodes with few links are likely to be under-represented.~\cite{tailgnn}. 
Such a degree-biased issue of GNNs should not be overlooked since the number of tail degree nodes greatly outnumbers the number of head degree nodes in real-world graphs as depicted in Figure~\ref{fig:figure1}(b). 

Few recent studies have attempted to address the above challenges.
Regarding the class long-tailed situation, most existing methods adopt an oversampling-based approach that interpolates nodes belonging to minority (tail) classes \cite{drgcn, imgagn, graphsmote}. 
Specifically, GraphSMOTE~\cite{graphsmote} adopts the synthetic minority oversampling algorithm \cite{smote} in the embedding space, and the generated nodes are connected to the existing nodes via an edge generator.
Most recently, GraphENS \cite{graphens} proposed an augmentation-based method that synthesizes an ego-network for nodes in the minority classes.
Unlike GraphSMOTE that relies only on the nodes in the minority classes, GraphENS additionally utilizes other classes' nodes for synthetic generation of nodes.
Moreover, regarding the degree long-tailed situation, recent studies aim at obtaining high-quality representations of tail degree nodes~\cite{metatail2vec, tailgnn}. Specifically, Tail-GNN~\cite{tailgnn} introduces the concept of neighborhood translation that aims to transfer the information in head degree nodes to tail degree nodes. 


\begin{figure}[t]
  \includegraphics[width=0.7\columnwidth]{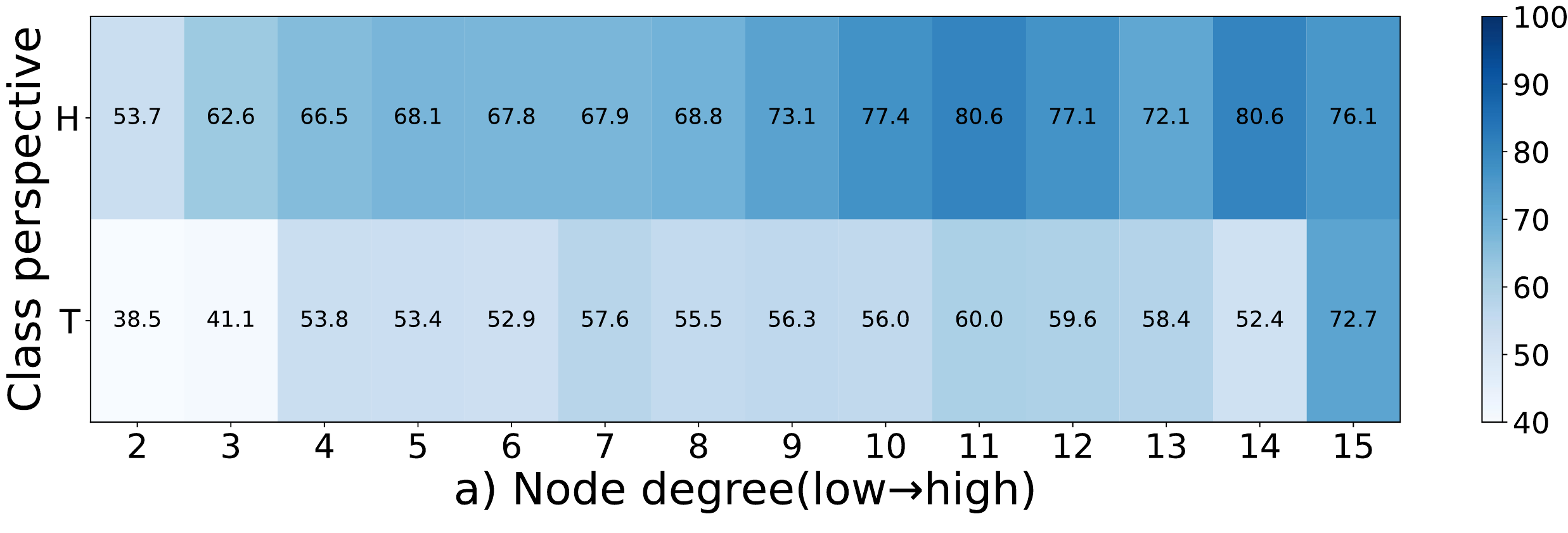}
  \includegraphics[width=0.7\columnwidth]{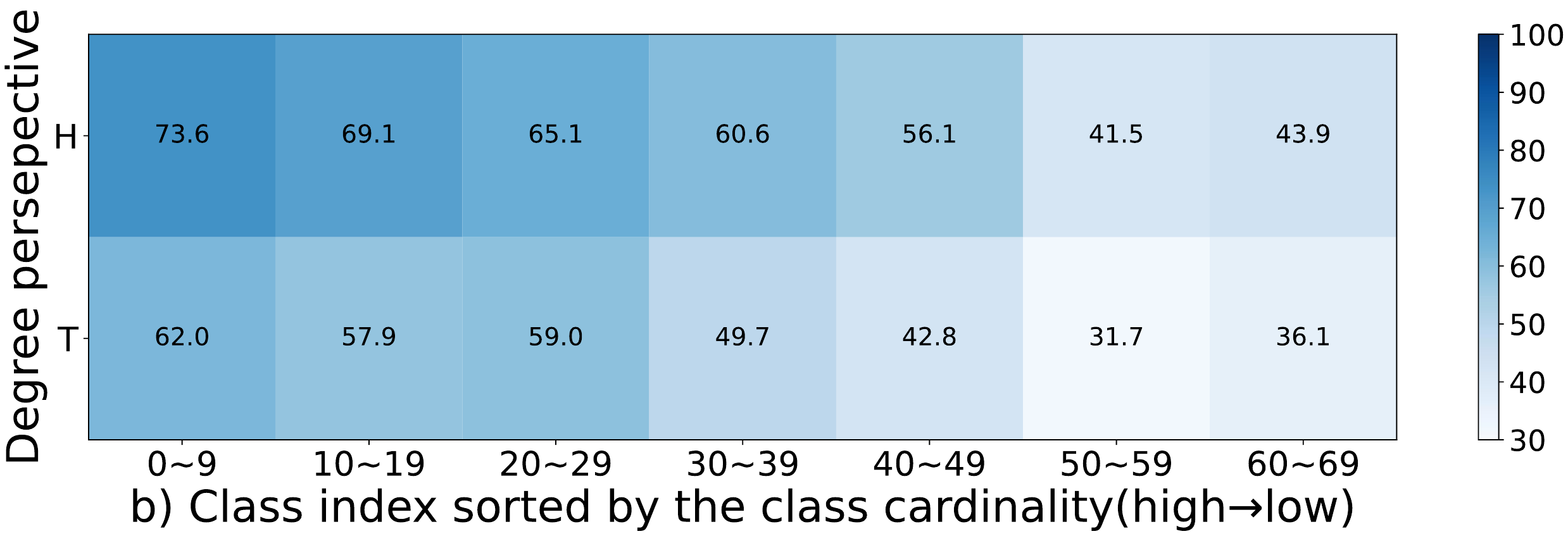}
    \vspace{-3ex}
  \caption{a) Class Head/Tail Separation with node degrees from low to high, and b) Degree Head/Tail Separation with class cardinality from high to low. Color indicates bAcc (\%) of node classification
  (Cora-Full is used).}
  \vspace{-5ex}
\label{fig:figure2}
\end{figure}

%

\looseness=-1
Despite the effectiveness of the aforementioned methods, they focus on only one of the two challenges even though jointly considering both challenges is crucial. 
To empirically demonstrate the importance of the joint consideration, we split the nodes in Cora and Cora-Full datasets considering both the class- and degree-longtailedness, and show the frequency with the node classification results in terms of balanced accuracy (i.e., bAcc) for each of the subsets in Figure~\ref{fig:figure1}(c). For example, HH indicates the nodes that jointly belong to Head class and Head degree. In Figure~\ref{fig:figure1}(c), we observe that the node classification performance on nodes that belong to Tail class (i.e., TH or TT) depends on whether the nodes belong to Head or Tail degree: HT<TH for Cora, while HT>TH for Cora-Full. Such a trend can be seen more clearly in Figure~\ref{fig:figure2}, where high degree nodes perform well even if they belong to a tail class (Figure~\ref{fig:figure2}(a)), and even high degree nodes do not always generalize well especially when the class cardinality is low (Figure~\ref{fig:figure2}(b)). Our observations motivate us to jointly consider both the class- and degree-longtailedness rather than considering them independently.

In this paper, we propose a novel framework for training GNNs, called \textsf{L}ong-\textsf{T}ail \textsf{E}xperts for \textsf{G}raphs (\proposed), which jointly considers the class long-tailedness, and the degree long-tailedness for node classification. 
The core idea is to assign an expert GNN model for each subset of nodes that are split in a balanced manner considering both the class and degree long-tailedness.
More precisely, we first split the nodes into two subsets considering the class distribution. i.e., Head class and Tail class. Then, we further split each subset considering the node degree distribution, which eventually results in four balanced subsets, i.e., 1) Head class \& Head degree (HH), 2) Head class \& Tail degree (HT), 3) Tail class \& Head degree (TH), and 4) Tail class \& Tail degree (TT).
After having trained an expert, i.e., a GNN-based classifier, for each balanced subset, we adopt knowledge distillation to obtain two class-wise students, i.e., Head class student and Tail class student, each of which is responsible for classifying nodes in Head class and Tail class, respectively. 
Lastly, 
we devise a class prototype-based inference that assigns each test node at the inference phase to either Head class or Tail class based on its similarity with class prototypes.
We demonstrate that~\proposed~outperforms a wide range of state-of-the-art methods in the node classification task evaluated on both manual and natural imbalanced graphs.
To the best of our knowledge, this is the first work that jointly considers the class long-tailedness and the degree long-tailedness on graphs. 

\section{PRELIMINARY}

\subsection{Graph Neural Networks}
\label{sec:2.1}
\looseness=-1
\noindent{\textbf{Graph.}} We denote $\mathcal{G}=(\mathcal{V},\mathcal{E},\mathbf{X})$ as a graph, where $\mathcal{V}=\{v_1,...,v_N\}$ represents the set of nodes, $\mathcal{E} \subseteq \mathcal{V} \times \mathcal{V}$ is the set of edges, and $\mathbf{X} \in \mathbb{R}^{N \times F}$ is the node feature matrix. $\mathbf{A} \in \mathbb{R}^{N \times N}$ is the adjacency matrix where $\mathbf{A}_{ij}=1$ if $(v_i, v_j) \in \mathcal{E}$ and $\mathbf{A}_{ij}=0$ otherwise. We denote $\mathcal{C}$ as the set of classes of nodes in $\mathcal{G}$.

\smallskip
\noindent{\textbf{Graph Neural Networks.}} Given an input node feature matrix $\mathbf{X}$ and the adjacency matrix $\mathbf{A}$, GNNs
update the representation of a node via aggregating its neighboring nodes. Formally, the general updating process of the $l$-th layer is as follows:
\vspace{-1ex}
\begin{equation}
\small
\mathbf{h}_{v}^{(l)}=\text{COMBINE}^{(l)}\left(\mathbf{h}_{v}^{(l-1)}, \text{AGGREGATE}^{(l-1)}\left(\left\{\mathbf{h}_{u}^{(l-1)}: u \in \mathcal{N}(v)\right\}\right)\right)
\end{equation}
where $\mathbf{h}_{v}^{(l)}\in\mathbb{R}^{D}$ denotes the representation of node $v \in \mathcal{V}$ at the $l$-th layer, and $\mathcal{N}(v)$ is a set of neighbors of node $v$.
Among many variants of GNNs, GCN \cite{gcn} is the most representative structure of GNN which is defined as follows: $\mathbf{H}^{(l)} = \sigma(\mathbf{\hat{D}}^{-1/2}\mathbf{\hat{A}}\mathbf{\hat{D}}^{-1/2}\mathbf{H}^{(l-1)}\mathbf{W}^{(l-1)})$,
where $\mathbf{H}^{(l)}\in\mathbb{R}^{N\times D}$ denotes the node embedding matrix at the $l$-th layer, $\hat{\mathbf{A}} = \mathbf{A} + \mathbf{I}$ is the adjacency matrix with self-loops, $\hat{\mathbf{D}} = \sum_{i}{\hat{\mathbf{A}}_i}$ is the degree matrix, $\mathbf{W}^{(l-1)}$ is the weight matrix at the $(l-1)$-th layer, $\mathbf{H}^{(0)}=\mathbf{X}$, and $\sigma$ is nonlinear activation such as ReLU. We adopt GCN as the backbone of our GNNs throughout the paper.

\subsection{Problem Statement}
\label{sec:2.2}
  Given a graph $\mathcal{G}=(\mathcal{V},\mathcal{E}, X)$ whose class and degree distributions of nodes are imbalanced, we aim to learn a GNN-based classifier that works well on both the head classes (i.e., classes with abundant samples) and the tail classes (i.e., classes with few samples), while also performing well considering the node degree perspective. 

\section{Proposed Framework: \proposed}
\begin{figure*}[t]
\centering
\includegraphics[width=1.4\columnwidth]{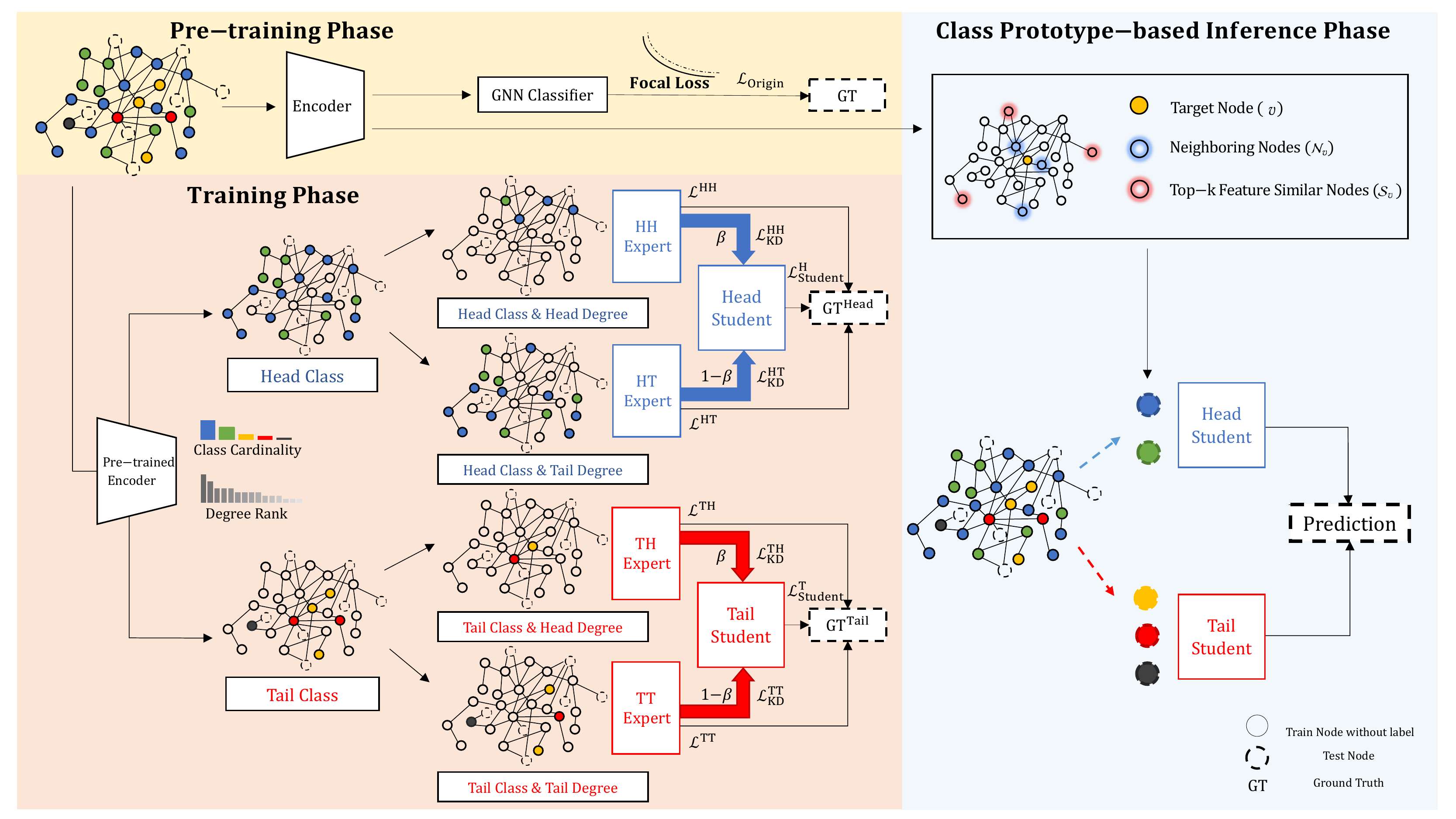}
\vspace{-2ex}
\caption{The overall architecture of~\proposed. Given a long-tailed graph in both class and degree perspective, we first obtain a pre-trained encoder at the pre-traing phase. At the training phase, we first split nodes of an imbalanced graph into balanced subsets, and then obtain experts specialized in its responsible subset followed by students that contain experts’ knowledge. Finally, at the class prototype-based inference phase, we assign each test node to a student and make the final prediction.}
\label{fig:figure3}
\vspace{-2ex}
\end{figure*}

\looseness=-1
In this section, we present a novel framework for training GNNs, called \textsf{L}ong-\textsf{T}ail \textsf{E}xperts for \textsf{G}raphs (\proposed), which jointly considers the class long-tailedness, and the degree long-tailedness for node classification.
We first explain how we train a pretrained encoder via focal loss (Sec 3.1), and how the original imbalanced graph is balanced (Sec 3.2). Then, we introduce long-tail experts that jointly consider the class and degree long-tailedness (Sec 3.3). Then, we demonstrate how to distill the experts' knowledge to head and tail students (Sec 3.4), followed by the description of the overall model training process (Sec 3.5). Lastly, we devise a class prototype-based inference (Sec 3.6). Figure~\ref{fig:figure3} shows the overall architecture of \proposed.


\subsection{Pre-training Phase}
\label{sec:3.1}
\looseness=-1
To obtain good initial node embeddings, we begin by pretraining a GCN encoder on the original graph to obtain node embeddings. The encoder is defined as: $\mathbf{H}^{\text{pre}} = \sigma(\mathbf{\hat{D}}^{-1/2}\mathbf{\hat{A}}\mathbf{\hat{D}}^{-1/2}\mathbf{X}\mathbf{W}^{\text{pre}})$,
where $\mathbf{X}\in\mathbb{R}^{N\times F}$ is the feature matrix, and $\mathbf{H}^{\text{pre}}\in\mathbb{R}^{N\times D}$ is the node embedding matrix generated by the encoder, which is passed to a GNN-based classifier to obtain class logits for nodes in the original imbalanced graph as follows: 
\begin{equation}
\small
\label{eqn:og_class}
\mathbf{P}^\text{og} = \text{softmax}(\mathbf{Z}^\text{og}), \;\; \mathbf{Z}^\text{og} = \sigma(\mathbf{\hat{D}}^{-1/2}\mathbf{\hat{A}}\mathbf{\hat{D}}^{-1/2}\mathbf{H}^{\text{pre}}\mathbf{W}^{og}_{\text{GNN}})\mathbf{W}^{og}_{\text{MLP}}
\end{equation}
where $\mathbf{W}^{og}_{\text{GNN}}\in\mathbb{R}^{D\times D}$ and $\mathbf{W}^{og}_{\text{MLP}}\in\mathbb{R}^{D\times\mathcal{|C|}}$ is a weight matrix for GNN and MLP, respectively, $\mathbf{Z}^\text{og}\in\mathbb{R}^{N\times\mathcal{|C|}}$ and $\mathbf{P}^\text{og}\in\mathbb{R}^{N\times\mathcal{|C|}}$ are the class logits and class probabilities for nodes, respectively.
However, directly training the encoder through the conventional cross-entropy loss on the imbalance data incurs predictions that are biased towards head classes. To prevent the bias, we adopt focal loss \cite{focal} that assigns higher weights to misclassified samples than to well-classified ones so as to focus on classifying the misclassified samples.
Specifically, considering the class imbalance situation in our case, we use $\alpha$-balanced variant of the focal loss: $FL(p_t) = -\alpha_{t} (1-p_t)^{\gamma}\log{p_t}$,
where given the predicted class probability $p$, $p_t=p$ for the ground truth class, otherwise $p_t=(1-p)$. $\alpha_t$ is a weighting factor, and $\gamma$ controls the shape of the curve, respectively.
The encoder is trained based on the focal loss as follows:
\begin{equation}
\small
\mathcal{L}_\text{Origin} = \sum_{v \in V} \sum_{c\in\mathcal{C}} FL(\mathbf{P}_v^\text{og}[c])
\end{equation}
where $\mathbf{P}^\text{og}_v$ is the $v$-th row of $\mathbf{P}^\text{og}$, and $\mathbf{P}_v^\text{og}[c]$ is the $c$-th element of $\mathbf{P}^\text{og}_v$. To summarize, we obtain $\mathbf{W}^{\text{pre}}$ as a result of pretraining, and this is used to initialize the encoders used in the training phase.


\subsection{Balancing the Node Distributions}
\label{sec:3.2}
Based on our observations in Figures \ref{fig:figure1} and \ref{fig:figure2}, we split the nodes in a graph in a balanced manner considering the long-tailedness in both the class distribution and the node degree distribution. More precisely, we first count the number of nodes in each class, and sort the classes according to the class cardinality. Then, the top-$p$\% classes are considered as the Head classes, whereas the remaining ($100-p$)\% classes are considered as the Tail classes, where $p$ is the hyperparameter that controls the separation. By doing so, we expect each subset of nodes to have a balanced class distribution of nodes. Then, we further split each subset considering the node degree distribution. Following recent methods~\cite{metatail2vec, tailgnn} that tackle the degree long-tailedness problem, we consider nodes with less than or equal to degree 5 as tail degree nodes, while the remaining nodes with degree greater than 5 are considered as head degree nodes. As a result, we obtain four subsets of nodes, i.e., 1) Head class \& Head degree (HH), 2) Head class \& Tail degree (HT), 3) Tail class \& Head degree (TH), and 4) Tail class \& Tail degree (TT). We later demonstrate in Table~\ref{tab:balance} that obtaining balanced subsets is crucial for the model performance.


\subsection{Long-Tail Experts}
\label{sec:3.3}
Now that we have four relatively balanced subsets of nodes, we assign a GNN-based expert to each subset. The core idea is to train four experts each of which is specialized in its subset at hand. 
The GNN-based expert is defined as follows:
\begin{equation}
\small
\label{eqn:logit}
    \mathbf{P}^{*} = \text{softmax}(\mathbf{Z}^{*}), \;\; \mathbf{Z}^{*} = \sigma(\mathbf{\hat{D}}^{-1/2}\mathbf{\hat{A}}\mathbf{\hat{D}}^{-1/2}\mathbf{H}^{\text{pre}}\mathbf{W}^{*}_{\text{GNN}}) \mathbf{W}^{*}_{\text{MLP}}
\end{equation}
where $*\in\{\text{HH},\text{HT},\text{TH},\text{TT}\}$, $\mathbf{Z}^*\in\mathbb{R}^{N\times |\mathcal{C}|}$ is the logit matrix, $\mathbf{P}^{*}\in\mathbb{R}^{N\times |\mathcal{C}|}$ is the predicted class probability matrix, and $\mathbf{W}_{\text{GNN}}^{*}\in\mathbb{R}^{D\times D}$ and $\mathbf{W}_{{\text{MLP}}}^{*}\in\mathbb{R}^{D\times |\mathcal{C}|}$ are learnable weight matrices for GNN and MLP, respectively. Here, based on our empirical finidings that head degree nodes perform better than tail degree nodes in both Head Class and Tail Class (Figure~\ref{fig:figure1}(c)), we obtain $\mathbf{W}^{HT}_{\text{GNN}}$ and $\mathbf{W}^{TT}_{\text{GNN}}$ via finetuning obtained $\mathbf{W}^{HH}_{\text{GNN}}$ and $\mathbf{W}^{TH}_{\text{GNN}}$, respectively.
Note that the node embedding matrix (i.e., $\mathbf{H}^{\text{pre}}$) is obtained via the GCN encoder pretrained in Sec.~\ref{sec:3.1}.

Given the predicted class probability distribution (i.e., $ \mathbf{P}^{*}$) for each subset as defined in Equation~\ref{eqn:logit}, we compute the cross-entropy loss as follows:

\begin{equation}
\small
\label{eq:loss}
\mathcal{L}^{*}_{\text{Expert}} = \sum_{v \in V^{*}} \sum_{c\in\mathcal{C}^{*}} CE(\mathbf{P}_v^{*}[c])
\end{equation}
where $V^*$ and $\mathcal{C}^*$ are set of nodes and classes that belong to $*\in\{\text{HH},\text{HT},\text{TH},\text{TT}\}$, and $CE(\cdot)$ denotes the cross-entropy loss\footnote{As each subset is relatively balanced, we use the cross-entropy loss for simplicity, but the focal loss can also be used instead.}.
More specifically, considering HH and HT experts are derived from the same Head class, $\mathcal{C}^{\text{HH}}(=\mathcal{C}^{\text{HT}})$ denotes the Head class (i.e., $\mathcal{C}^{\text{H}}$). Likewise, $\mathcal{C}^{\text{TH}}(=\mathcal{C}^{\text{TT}})$ denotes the Tail class (i.e., $\mathcal{C}^{\text{T}}$).
Lastly, we add all the losses computed in Equation~\ref{eq:loss} as follows:

\begin{equation}
\small
\mathcal{L}_{\text{Expert}}=\sum_{* \in \{\text{HH},\text{HT},\text{TH},\text{TT}\}} \mathcal{L}^{*}_{\text{Expert}}
\end{equation}
Such balanced training is inspired by \cite{lfme} whose argument is that the model trained with fewer samples on a balanced dataset tends to perform better compared with the model trained with more samples on a long-tailed dataset. Although optimizing the above loss provides accurate node classification results on classes and degrees for which each expert is responsible, the remaining challenge is how to leverage the experts' knowledge so as to obtain the final node classification result, which is our ultimate goal.

\subsection{Distilling Experts' Knowledge to Students}
\label{sec:3.4}
\looseness=-1

It is important to note that knowledge obtained in the previous step may sometimes be noisy when the number of allocated nodes for a certain expert is not enough. To alleviate such circumstances and further leverage the experts' knowledge, we introduce two students, i.e., Head class student and Tail class student, each of which is responsible for classifying nodes that belong to the Head class and the Tail class. Then, we let them acquire knowledge from the experts. The GNN-based student is defined as follows:
\begin{equation}
\small
\label{eqn:student}
    \mathbf{P}^{\star} = \text{softmax}(\mathbf{Z}^{\star}), \;\; \mathbf{Z}^{\star} = \sigma(\mathbf{\hat{D}}^{-1/2}\mathbf{\hat{A}}\mathbf{\hat{D}}^{-1/2}\mathbf{H}^{\text{pre}}\mathbf{W}^{\star}_{\text{GNN}}) \mathbf{W}^{\star}_{\text{MLP}}
\end{equation}
where ${\star}\in\{\text{H},\text{T}\}$, $\mathbf{Z}^{\star}$ is the logit matrix, $\mathbf{P}^{\star}$ is the predicted class probability matrix, and $\mathbf{W}_{\text{GNN}}^{\star}\in\mathbb{R}^{D\times D}$ and $\mathbf{W}_{{\text{MLP}}}^{\star}$ are learnable weight matrices for GNN and MLP, respectively.
Having defined the two students as above, 
we adopt the knowledge distillation technique~\cite{hinton2015distilling} to distill knowledge from the experts learned in the previous step. More precisely, the Head class student learns from HH expert and HT expert and the Tail class student learns from TH expert and TT expert. 
To this end, we compute the KL-divergence between the probability distributions obtained from the logits of a student and its associated experts. As for the Head class student, we perform knowledge distillation as:
\begin{equation}
\small
    \mathcal{L}_{\text{KD}}^{\text{HH}}= D_\text{KL}[\mathbf{P}^{\text{HH}}\Vert {\mathbf{P}}^{\text{H}}], \;\;
    \mathcal{L}_{\text{KD}}^{\text{HT}}= D_\text{KL}[\mathbf{P}^{\text{HT}}\Vert {\mathbf{P}}^{\text{H}}]
    \label{eq:distill_H}
\end{equation}
where $D_\text{KL}[\cdot \Vert \cdot]$ is the KL-divergence, $\mathcal{L}_{\text{KD}}^{\text{HH}}$ and $\mathcal{L}_{\text{KD}}^{\text{HT}}$ are the knowledge distillation losses for HH expert and HT expert, respectively, and $\mathbf{P}^{\text{HH}}$, $\mathbf{P}^{\text{HT}}$, ${\mathbf{P}}^{\text{H}}$ are the predicted class probability distributions for HH expert, HT expert, and Head class student, respectively.
Moreover, the knowledge distillation losses for the Tail class student can be similarly defined as follows:

\begin{equation}
\small
    \mathcal{L}_{\text{KD}}^{\text{TH}}= D_\text{KL}[\mathbf{P}^{\text{TH}}\Vert {\mathbf{P}}^{\text{T}}],   \;\; \mathcal{L}_{\text{KD}}^{\text{TT}}= D_\text{KL}[\mathbf{P}^{\text{TT}}\Vert {\mathbf{P}}^{\text{T}}]
    \label{eq:distill_T}
\end{equation}
where $\mathcal{L}_{\text{KD}}^{\text{TH}}$ and $\mathcal{L}_{\text{KD}}^{\text{TT}}$ are the knowledge distillation losses for TH expert and TT expert, respectively, $\mathbf{P}^{\text{TH}}$, $\mathbf{P}^{\text{TT}}$, ${\mathbf{P}}^{\text{T}}$ are the predicted class probability distributions for TH expert, TT expert, and Tail class student, respectively. In summary, by minimizing the distillation losses defined in Equations~\ref{eq:distill_H} and~\ref{eq:distill_T}, the experts' knowledge can be distilled to students.

\smallskip
\looseness=-1
\noindent\textbf{Distilling Knowledge from Head- to Tail-degree expert.}
However, it is important to note that since the two subsets of nodes that are assigned to the same student differ in the node degree, the performance of the expert that is assigned to the subset with high degree nodes is superior to its counterpart. For example, in the case of the Head class student, which is associated with HH subset and HT subset, since nodes in HH have higher degree than nodes in HT, HH expert would perform better than HT expert. In fact, this can also be observed from our empirical analysis shown in Figure \ref{fig:figure1}(c) (i.e., HH$>$HT and TH$>$TT in terms of bAcc). Hence, for students, the knowledge contained in a Tail degree expert (e.g., HT) is more difficult to acquire than that contained in a Head degree expert (e.g., HH). 
Based on the above intuition, we employ curriculum learning while distilling the knowledge of Head degree expert and Tail degree expert to students. More precisely, we want students to learn more from Head degree experts in the early stage of training, while in the later stage we want students to learn more from Tail degree experts. The loss for Head class student is defined as:
\begin{equation}
\small
\label{eqn:Head class student}
    \mathcal{L}_{\text{Student}}^{\text{H}}= \beta \mathcal{L}_{\text{KD}}^{\text{HH}}+ (1-\beta) \mathcal{L}_{\text{KD}}^{\text{HT}} 
    \vspace{-1ex}
\end{equation}
where $\beta$ is a scheduling hyperparameter that progressively becomes smaller as training proceeds. Moreover, the loss for the Tail class student is defined in a similar way:
\begin{equation}
\small
\label{eqn:Tail class student}
\mathcal{L}_{\text{Student}}^{\text{T}}= \beta \mathcal{L}_{\text{KD}}^{\text{TH}}+ (1-\beta) \mathcal{L}_{\text{KD}}^{\text{TT}}
\vspace{-1ex}
\end{equation}
Note that we adopt the convex scheduler, i.e., $\beta = \cos{\frac{e\pi}{E^2}}$, where $E$ and $e$ denote the total epochs and the current epoch, respectively. 
Besides the knowledge distillation losses (i.e., $\mathcal{L}_{\text{Student}}^{\text{H}}$ and $\mathcal{L}_{\text{Student}}^{\text{T}}$), we compute cross-entropy loss for Head class student and Tail class students as follows:
\begin{equation}
\small
\mathcal{L}_{\text{CE}} = \sum_{\star \in \{\text{H},\text{T}\}} \mathcal{L}^{\star}_{\text{CE}}, \;\; \mathcal{L}^{\star}_{\text{CE}} = \sum_{v \in V^{\star}} \sum_{c\in\mathcal{C}^{\star}} CE(\mathbf{P}_v^{\star}[c])
\end{equation}
Putting them all together, the final knowledge distillation loss is defined as follows:
\begin{equation}
\small
\mathcal{L}_{\text{Student}}= \mathcal{L}_{\text{Student}}^{\text{H}} + \mathcal{L}_{\text{Student}}^{\text{T}} + \mathcal{L}_{\text{CE}}
\end{equation}
By optimizing the above loss, each student not only gains knowledge regarding the class information, but also the degree information,
thereby being able to better cope with both the class-longtailedness and the degree-longtailedness of nodes in a graph.

\subsection{Model Training}
Considering the losses for the experts and the students, the final objective function of~\proposed~is defined as follows:
\begin{equation}
\small
\mathcal{L}_{\text{Final}} =  \underbrace{\mathcal{L}_{\text{Expert}}}_{\text{Sec.~\ref{sec:3.3}}} +  \underbrace{\mathcal{L}_{\text{Student}}}_{\text{Sec.~\ref{sec:3.4}}}
\end{equation}
It is important to note that the class-wise students begin to acquire knowledge from the experts once the experts are trained until convergence (in terms of validation macro-f1). In other words, $\mathcal{L}_{\text{Expert}}$ is first trained until convergence, then $\mathcal{L}_{\text{Student}}$ is trained while having the parameters involved in $\mathcal{L}_{\text{Expert}}$ fixed.

\vspace{-1ex}
\subsection{Class Prototype-based Inference Phase}
\label{sec:3.6}
\looseness=-1
Since~\proposed~performs node classification based on two class-wise students, i.e., Head class student and Tail class student, the main challenge at the inference phase is how to determine whether a test node should be sent to the Head class student or the Tail class student.
This is crucial as correctly classifying a test node would be impossible once the node is assigned to an incorrect student.
To this end, we devise a prototype-based inference, whose main idea is to assign each test node to one of the students based on its similarity with class prototypes. That is, for a given test node, we find the class whose prototype is the most similar to the test node, and then assign the test node to the corresponding student.
The most naive way of computing class prototypes would be to compute the average of the pretrained embeddings of labeled nodes in the training data that belong to each class. More precisely, let $\mathcal{V}^c_\text{train}$ denote the set of training nodes that belong to class $c\in\mathcal{C}$, where $v^c_\text{train}$ is a node that belongs to $\mathcal{V}^c_\text{train}$. Then, the class prototype for class $c\in\mathcal{C}$ (i.e., $\mathbf{p}^c\in\mathbb{R}^D$) is obtained by computing the average of embeddings of $v^c_\text{train}\in\mathcal{V}^c_\text{train}$ as follows:
\begin{equation}
\small
\mathbf{p}^c=\frac{1}{|\mathcal{V}^c_\text{train}|} \sum_{v^c_\text{train}\in\mathcal{V}^c_\text{train}}\mathbf{H}^{\text{pre}}_{v^c_\text{train}}    
\label{eq:proto}
\end{equation}
where $\mathbf{H}^{\text{pre}}_{v^c_\text{train}}\in\mathbb{R}^D$ is the pretrained embedding of $v^c_\text{train}$ with a slight abuse of notation.
After obtaining $\mathbf{p}^c$ for all $c\in\mathcal{C}$, we compute the similarity between the node embedding of a given test node $v_\text{test}$ (i.e.,$\mathbf{H}^{\text{pre}}_{v_\text{test}}$) and the class prototypes $\mathbf{p}^c$ for all $c\in\mathcal{C}$ to determine the class $c$ whose similarity is the maximum:
\begin{equation}
\vspace{-1ex}
\small
    c^*=\argmax_{c}\textsf{sim}(\mathbf{p}_c,\mathbf{H}^{\text{pre}}_{v_\text{test}}),\; \forall c \in \mathcal{C}
    \label{eq:c}
\end{equation}
where $\textsf{sim}()$ is the cosine similarity. Having found $c^*$ from Equation~\ref{eq:c}, we assign $v_\text{test}$ to the Head class student if $c^*\in\mathcal{C}^{\text{H}}$, and to the Tail class student if $c^*\in\mathcal{C}^{\text{T}}$. In other words, if a given test node $v_\text{test}$ is assigned to the Head class student, the final predicted class is obtained as: $\hat{\textbf{Y}}_{v_\text{test}} = \argmax_{c} \mathbf{Z}^\text{H}_{v_\text{test}}$,
where $\mathbf{Z}^\text{H}_{v_\text{test}}\in\mathbb{R}^{|\mathcal{C}^{\text{H}}|}$ is the class logit vector of $v_\text{test}$ obtained from the Head class student, and $\hat{\textbf{Y}}_{v_\text{test}}$ is the final predicted class for $v_\text{test}$. The case when $v_\text{test}$ is assigned to the Tail class student is similarly defined.

\smallskip
\noindent\textbf{Expanding Candidates for Class Prototype Computation.}
However, we argue that as the class distribution is usually imbalanced, the number of nodes used to compute the class prototypes would vary greatly. For example, in an extreme case, a class with a single labeled node would have to rely on that single node to compute the class prototype.
This implies that some of the class prototypes are of lower-quality than others.

Hence, in order to expand the candidate nodes for the class prototype computation, rather than solely relying on the labeled nodes (i.e., $\mathcal{V}^c_\text{train}$), we additionally consider their neighboring nodes, i.e., nodes that are linked. The intuition is that as most real-world graphs follow the homophily assumption~\cite{mcpherson2001birds,newman2018networks}, which implies that linked nodes often belong to the same class, we find that incorporating the neighboring nodes as candidate nodes for the class prototype computation (even though they are not labeled) help compute higher-quality class prototypes. More formally, we denote $\mathcal{N}_{v_\text{train}^c}$ as the set of neighboring nodes of $v_\text{train}^c\in\mathcal{V}^c_\text{train}$.
Despite the homophily assumption, not all the neighboring nodes share the same label as the the target node $v_\text{train}^c$, which may lead to noise in the class prototype computation. Hence, we leverage the pre-computed class probabilities ($\mathbf{P}^\text{og}$ in Equation~\ref{eqn:og_class}) to select neighboring nodes with high confidence of belonging to the same class as the target node. Moreover, as the node degree also exhibits a long-tail distribution, most nodes only have few neighboring nodes. In this regard, we further expand the candidate nodes to also incorporate nodes with similar features (i.e., $\mathbf{X}$). More precisely, given $v_\text{train}^c$, we compute its top-$k$ similar nodes in terms of the node feature $\mathbf{X}$, and denote by $\mathcal{S}_{v^c_\text{train}}$. Likewise, we select similar nodes with high confidence of belonging to the same class as $v_\text{train}^c$. Then, we expand the candidate nodes for the class prototype computation by also considering $\mathcal{S}_{v^c_\text{train}}$ to obtain: $\{\mathcal{V}^c_\text{train} \cup \mathcal{N}^c_{\text{train}} \cup \mathcal{S}^c_{\text{train}}\}$, where $\mathcal{S}_\text{train}^c=\bigcup\limits_{v_\text{train}^c\in\mathcal{V}^c_\text{train}}\mathcal{S}_{v_\text{train}^c}$. We empirically observed that the model performs the best when we incorporate neighboring nodes first and then supplement the remaining candidates with similar nodes. Here, the number of incorporating nodes for candidates can be set as the mean or max of the number of training samples per class. Ablation studies in this regard will be later shown in Figure~\ref{fig:figure5}. It is important to note that as pretrained node embeddings (i.e., $\mathbf{H}^{\text{pre}}$) and predicted class probabilities (i.e., $\mathbf{P}^\text{og}$) are used to compute the class prototypes, all the class prototypes can be computed in advance, which makes the inference scalable in practice.

\section{Experiments}


\subparagraph{\textbf{Datasets.}} To perform node classification, we evaluate our proposed framework on Cora, CiteSeer~\cite{citation} and Cora-Full~\cite{cora-full} datasets. Following the evaluation protocol of recently proposed methods~\cite{graphsmote,graphens}, we manually preprocess the imbalance ratio of Cora and Citeseer datasets, while Cora-Full dataset is used per se to evaluate the models on a natural dataset without any manual preprocessing step.
To be specific, 
GraphSMOTE~\cite{graphsmote} assumes that all majority classes (i.e., Head classes) have 20 training nodes and minority classes (i.e., Tail classes) have 20 $\times$ \textit{imbalance\_ratio} nodes for each class, where the \textit{imbalance\_ratio} is the ratio between the number of samples in the minority class and the number of samples in the majority class.
Here, we consider harsh imbalance situations and set the \textit{imbalance\_ratio} to 10\% and 5\% for each imbalanced dataset. Besides conventional settings~\cite{graphsmote,graphens}, we also evaluate under the circumstances where the number of imbalanced class varies. Specifically, we evaluate models when the number of minority classes is either 3 or 5. We also follow the protocol of GraphENS \cite{graphens} in which a long-tailed versions of Cora and CiteSeer, i.e., Cora-LT and CiteSeer-LT, are generated by sorting the classes in descending order and removing nodes with low-degrees in minority classes until a certain \textit{imbalance\_ratio} (e.g., 1\%) is met. For the validation and test sets, we follow GraphSMOTE~\cite{graphsmote} and fix the same number of validation and test samples for each class. Furthermore, to verify whether~\proposed~works well not only on manual imbalanced datasets, but also on a natural imbalance dataset, we evaluate our model on Cora-Full dataset whose number of classes is large (i.e., $70$) and at the same time the \textit{imbalance\_ratio} is harsh (i.e., $\frac{1}{92}\times100=1.1\%$).
To sum up, ~\proposed~is evaluated based on the manual imbalanced settings proposed by GraphSMOTE~\cite{graphsmote} and GraphENS~\cite{graphens}, and an additional natural dataset, Cora-Full~\cite{cora-full}. For more details regarding the datasets, refer to Table~\ref{tab:table1} and Table~\ref{tab:table2}.

\begin{table}[t]
\small
\caption{\label{tab:table1}Statistics of datasets.}
\vspace{-3ex}
\resizebox{0.75\columnwidth}{!}{
\begin{tabular}{c|cccc}
\hline
\textbf{Dataset}   & \textbf{\#Nodes} & \textbf{\#Edges} & \textbf{\#Features} & \textbf{\#Classes} \\ \hline \hline
\textbf{Cora}      & 2,708         & 5,429         & 1,433             & 7              \\ \hline
\textbf{CiteSeer}  & 3,327         & 4,732         & 3,703             & 6              \\ \hline
\textbf{Cora-Full} & 19,793        & 146,635       & 8,710             & 70             \\ \hline
\end{tabular}
}
\vspace{-1ex}

\end{table}

\begin{table}[t]

\caption{\label{tab:table2}Summary of imbalanced settings. $\text{L}_{i}(\%)$ denotes the ratio of each class sorted by the number of samples, while $\text{L}_{i}(\%)$ of Cora-Full denotes the ratio of the $i$-th bin where each bin contains ten classes.}
\vspace{-2ex}
\resizebox{0.95\columnwidth}{!}{
\begin{tabular}{c|c|c|ccccccc}
\hline
\textbf{Dataset}                   & \textbf{Imb. class} & \textbf{Imb. ratio} & \textbf{$\text{L}_{0}$} & \textbf{$\text{L}_{1}$} & \textbf{$\text{L}_{2}$} & \textbf{$\text{L}_{3}$} & \textbf{$\text{L}_{4}$} & \textbf{$\text{L}_{5}$} & \textbf{$\text{L}_{6}$} \\ \hline \hline
\multirow{5}{*}{\textbf{Cora}}     & \multirow{2}{*}{3}  & 10\%                & 23.3                    & 23.3                    & 23.3                    & 23.3                    & 2.4                     & 2.4                     & 2.4                     \\ \cline{3-3}
                                   &                     & 5\%                 & 24.1                    & 24.1                    & 24.1                    & 24.1                    & 1.2                     & 1.2                     & 1.2                     \\ \cline{2-10} 
                                   & \multirow{2}{*}{5}  & 10\%                & 40.0                    & 40.0                    & 4.0                     & 4.0                     & 4.0                     & 4.0                     & 4.0                     \\ \cline{3-3}
                                   &                     & 5\%                 & 44.4                    & 44.4                    & 2.2                     & 2.2                     & 2.2                     & 2.2                     & 2.2                     \\ \cline{2-10} 
                                   & LT                  & 1\%                 & 54.0                    & 25.0                    & 11.6                    & 5.4                     & 2.4                     & 1.2                     & 0.5                     \\ \hline
\multirow{5}{*}{\textbf{CiteSeer}} & \multirow{2}{*}{3}  & 10\%                & 30.3                    & 30.3                    & 30.3                    & 3.0                     & 3.0                     & 3.0                     & -                       \\ \cline{3-3}
                                   &                     & 5\%                 & 31.7                    & 31.7                    & 31.7                    & 1.6                     & 1.6                     & 1.6                     & -                       \\ \cline{2-10} 
                                   & \multirow{2}{*}{5}  & 10\%                & 66.7                    & 6.7                     & 6.7                     & 6.7                     & 6.7                     & 6.7                     & -                       \\ \cline{3-3}
                                   &                     & 5\%                 & 80.0                    & 4.0                     & 4.0                     & 4.0                     & 4.0                     & 4.0                     & -                       \\ \cline{2-10} 
                                   & LT                  & 1\%                 & 60.7                    & 24.1                    & 9.5                     & 3.8                     & 1.5                     & 0.5                     & -                       \\ \hline
\textbf{Cora-Full}                 & -                   & 1.1\%               & 34.0                    & 18.9                    & 14.1                    & 10.9                    & 6.9                     & 4.8                     & 2.6                     \\ \hline
\end{tabular}
}
\vspace{-3ex}
\end{table}

\smallskip
\subparagraph{\textbf{Compared Methods.}} As~\proposed~uses the GNN backbone (i.e., GCN encoder~\cite{gcn}) and jointly considers the class long-tailedness and degree long-tailedness, we compare our framework with the following methods that address imbalanced datasets: \\
(1) Classical approaches for imbalanced learning

$\bullet$ \hangindent=0.7cm \textbf{Over-sampling}: It repeatedly samples nodes that belong to the minority classes. Here, we duplicate the number of nodes in tail classes and obtain a new adjacency matrix containing oversampled node's connectivity.

$\bullet$ \hangindent=0.7cm \textbf{Re-weight} \cite{reweight}: In terms of cost-sensitive approach, we impose a penalty to tail nodes that belong to the minority classes to remedy the domination of the nodes that belong to the majority classes.

$\bullet$ \hangindent=0.7cm \textbf{SMOTE} \cite{smote}: It performs synthetic minority oversampling to generate nodes that belong to minority classes in raw feature space by interpolating a node with its k-NN based neighbors that belong to the same class. For the generated node, we assign the same edges with its target neighbor's edges.

$\bullet$ \hangindent=0.7cm \textbf{Embed-SMOTE} \cite{embed-smote}: It performs SMOTE in the embedding space instead of feature space. We interpolate nodes that pass the last GNN layer, and thus we do not need to generate edges. 

\noindent (2) GNN-based imbalanced learning for class-longtailedness.

$\bullet$ \hangindent=0.7cm \textbf{GraphSMOTE} \cite{graphsmote}: It adopts SMOTE~\cite{smote} on node embeddings 
and connects the generated nodes via an edge generator. It has two versions depending on whether the predicted edges are discrete or continuous. We consider both versions as well as their pre-trained versions.

$\bullet$ \hangindent=0.7cm \textbf{GraphENS} \cite{graphens}: It is the state-of-the-art class-imbalanced learning method on graphs, which is based on augmentations of one-hop network of a minor node with that of the target node. It utilizes feature saliency information of the target node to prevent a generation of noisy samples.

\noindent (3) GNN-based imbalanced learning for degree-longtailedness.
    
$\bullet$ \hangindent=0.7cm \textbf{Tail-GNN} \cite{tailgnn}: It is the most recent work that learns embeddings for tail nodes in the degree perspective by transferring information from high-degree nodes to low-degree nodes.

\smallskip
\subparagraph{\textbf{Experimental Settings and Evaluation Metrics.}} For evaluations of~\proposed, we perform node classification under various manual imbalance settings, and a natural setting. Regarding the data splits, we follow the protocol of GraphSMOTE \cite{graphsmote} and GraphENS \cite{graphens} for the manual imbalance settings, and 1:1:8 (train/valid/test) splits for the natural setting. As for the evaluation metrics, besides conventional accuracy (Acc), we use several metrics widely used for imbalanced classification problems (i.e., balanced accuracy (bAcc), Macro-F1 and Geometric Means (G-Means) \cite{gmeans}) and report them along with standard deviations for 3 repetitions with different random seeds.


\smallskip
\looseness=-1
\noindent\textbf{Implementation Details. }
For all models, we use the Adam~\cite{adam} optimizer where the learning rate is 0.01 with weight deacy 5$e$-4, and hidden dimension is set as 64. For the oversampling-based baselines, we set the number of imbalanced classes to be the same as the number of tail classes in the manual imbalance settings, and otherwise selected in \{1,2,3,4,5\} for the natural imbalance dataset (i.e., Cora-Full). The upsampling ratio is fixed to 1.0 (i.e., oversamples the same amount of samples in each tail class) since it generally performs well on various settings, and we find that the ratio greater than 2.0 tends to give noise on the original dataset. For GraphSMOTE \cite{graphsmote}, we select the pretrain epoch in \{50, 100\} and set the reconstruction constant to 1$e$-6 as in the original paper. For GraphENS \cite{graphens} and Tail-GNN \cite{tailgnn}, we adopt their best hyperparameter settings reported in the papers. 
For~\proposed~, the class separation rule is selected in \{1/9,2/8,3/7,4/6,5/5,6/4,7/3,8/2,9/1\}, and regarding the hyperparameters for the focal loss, we select $\gamma$ in \{0,1,2\}, and $\alpha$ is selected in \{0.1,0.2,0.3,0.4,0.5,0.6,0.7,0.8,0.9\} or set to the inverse class frequency. Moreover, all the models are trained until their convergence with the maximum epoch set to 10000, and evaluated with their best model based on their validation Macro-F1 scores. 
We adopt the edge decoder proposed in GraphSMOTE~\cite{graphsmote}, which turns out to be helpful for the model training especially when the class imbalance is severe.
All the experiments are conducted on an RTX3090 (24GB).

\vspace{-1ex}

\begin{table*}[t]
\centering
\caption{\label{tab:table3}The overall performance on \textit{manual imbalanced} datasets (\textit{Imbalance\_ratio: 10\%, 5\%}). Note that in the manual imbalanced settings, the balanced accuracy is equivalent to accuracy, since the number of test nodes is same in each class.}
\vspace{-1ex}
\resizebox{0.8\textwidth}{!}{%
\begin{tabular}{cl|cccccc|cccccc}
\hline
\multicolumn{2}{c|}{\multirow{3}{*}{\textbf{Method}}}          & \multicolumn{6}{c|}{\textbf{Imb. class num: 3}}                                                                                                               & \multicolumn{6}{c}{\textbf{Imb. class num: 5}}                                                                                                                \\ \cline{3-14} 
\multicolumn{2}{c|}{}                                          & \multicolumn{3}{c|}{\textit{Imbalance\_ratio: 10\%}}                                                    & \multicolumn{3}{c|}{\textit{Imbalance\_ratio: 5\%}}                                & \multicolumn{3}{c|}{\textit{Imbalance\_ratio: 10\%}}                                                    & \multicolumn{3}{c}{\textit{Imbalance\_ratio: 5\%}}                                 \\ \cline{3-14} 
\multicolumn{2}{c|}{}                                          & bAcc.                & Macro-F1             & \multicolumn{1}{c|}{G-Means}              & bAcc.                & Macro-F1             & G-Means              & bAcc.                & Macro-F1             & \multicolumn{1}{c|}{G-Means}              & bAcc.                & Macro-F1             & G-Means              \\ \hline \hline
\multirow{12}{*}{\rotatebox{90}{Cora}}     & Origin            & 68.8\small{$\pm$4.0} & 67.6\small{$\pm$5.0} & \multicolumn{1}{c|}{80.8\small{$\pm$2.6}} & 60.0\small{$\pm$0.4} & 56.6\small{$\pm$0.7} & 74.8\small{$\pm$0.3} & 64.9\small{$\pm$6.2} & 64.7\small{$\pm$5.7} & \multicolumn{1}{c|}{78.1\small{$\pm$4.2}} & 55.1\small{$\pm$2.6} & 51.4\small{$\pm$2.4} & 71.4\small{$\pm$1.8} \\
                                           & Over-sampling     & 65.6\small{$\pm$4.3} & 63.4\small{$\pm$5.6} & \multicolumn{1}{c|}{78.6\small{$\pm$2.9}} & 59.0\small{$\pm$2.2} & 53.9\small{$\pm$2.6} & 74.2\small{$\pm$1.5} & 58.9\small{$\pm$5.6} & 56.9\small{$\pm$7.6} & \multicolumn{1}{c|}{74.0\small{$\pm$3.9}} & 49.1\small{$\pm$4.1} & 45.6\small{$\pm$5.4} & 67.0\small{$\pm$3.1} \\
                                           & Re-weight         & 70.6\small{$\pm$4.3} & 69.9\small{$\pm$5.1} & \multicolumn{1}{c|}{81.9\small{$\pm$2.8}} & 60.8\small{$\pm$1.8} & 56.7\small{$\pm$2.4} & 75.4\small{$\pm$1.3} & 65.2\small{$\pm$7.6} & 65.0\small{$\pm$8.2} & \multicolumn{1}{c|}{78.3\small{$\pm$5.2}} & 57.9\small{$\pm$4.3} & 54.8\small{$\pm$5.5} & 73.3\small{$\pm$3.0} \\
                                           & SMOTE             & 65.1\small{$\pm$4.0} & 62.3\small{$\pm$5.1} & \multicolumn{1}{c|}{78.3\small{$\pm$2.7}} & 59.0\small{$\pm$2.2} & 53.9\small{$\pm$2.6} & 74.2\small{$\pm$1.5} & 60.3\small{$\pm$7.6} & 58.7\small{$\pm$8.9} & \multicolumn{1}{c|}{74.9\small{$\pm$5.3}} & 49.1\small{$\pm$4.1} & 45.6\small{$\pm$5.4} & 67.0\small{$\pm$3.1} \\
                                           & Embed-SMOTE       & 61.0\small{$\pm$3.6} & 58.0\small{$\pm$5.5} & \multicolumn{1}{c|}{75.5\small{$\pm$2.5}} & 55.5\small{$\pm$2.1} & 50.0\small{$\pm$3.1} & 71.7\small{$\pm$1.5} & 53.2\small{$\pm$5.2} & 50.9\small{$\pm$6.4} & \multicolumn{1}{c|}{70.0\small{$\pm$3.8}} & 40.7\small{$\pm$2.8} & 36.5\small{$\pm$3.0} & 60.5\small{$\pm$2.2} \\
                                           & $\text{GraphSMOTE}_{T}$    & 70.0\small{$\pm$3.4} & 68.6\small{$\pm$4.9} & \multicolumn{1}{c|}{81.6\small{$\pm$2.2}} & 62.5\small{$\pm$1.8} & 58.7\small{$\pm$2.1} & 76.5\small{$\pm$1.2} & 66.3\small{$\pm$6.6} & 65.3\small{$\pm$7.7} & \multicolumn{1}{c|}{79.0\small{$\pm$4.5}} & 55.8\small{$\pm$5.6} & 52.4\small{$\pm$4.7} & 71.9\small{$\pm$4.0} \\
                                           & $\text{GraphSMOTE}_{O}$    & 67.3\small{$\pm$3.8} & 65.5\small{$\pm$5.0} & \multicolumn{1}{c|}{79.7\small{$\pm$2.5}} & 61.0\small{$\pm$0.6} & 58.4\small{$\pm$0.7} & 75.5\small{$\pm$0.4} & 62.8\small{$\pm$6.1} & 62.0\small{$\pm$6.1} & \multicolumn{1}{c|}{76.7\small{$\pm$4.2}} & 59.6\small{$\pm$4.3} & 55.9\small{$\pm$4.2} & 74.5\small{$\pm$3.0} \\
                                           & $\text{GraphSMOTE}_{preT}$ & 70.6\small{$\pm$6.3} & 68.9\small{$\pm$7.9} & \multicolumn{1}{c|}{81.9\small{$\pm$4.1}} & 67.5\small{$\pm$5.7} & 64.5\small{$\pm$8.0} & 79.9\small{$\pm$3.8} & 69.4\small{$\pm$5.2} & 68.3\small{$\pm$5.2} & \multicolumn{1}{c|}{81.1\small{$\pm$3.4}} & 66.0\small{$\pm$6.0} & 63.4\small{$\pm$6.5} & 78.8\small{$\pm$4.0} \\
                                           & $\text{GraphSMOTE}_{preO}$ & 69.8\small{$\pm$5.7} & 67.9\small{$\pm$7.1} & \multicolumn{1}{c|}{81.4\small{$\pm$3.7}} & 67.2\small{$\pm$5.2} & 64.0\small{$\pm$7.3} & 79.6\small{$\pm$3.5} & 69.1\small{$\pm$7.7} & 67.8\small{$\pm$7.9} & \multicolumn{1}{c|}{80.9\small{$\pm$5.1}} & 66.8\small{$\pm$4.9} & 64.6\small{$\pm$4.9} & 79.4\small{$\pm$3.2} \\
                                           & GraphENS          & 59.3\small{$\pm$7.0} & 55.4\small{$\pm$10.6} & \multicolumn{1}{c|}{74.2\small{$\pm$4.9}} & 55.1\small{$\pm$4.9} & 48.1\small{$\pm$7.9} & 71.3\small{$\pm$3.5} & 44.3\small{$\pm$6.5} & 41.0\small{$\pm$7.0} & \multicolumn{1}{c|}{63.3\small{$\pm$5.0}} & 36.1\small{$\pm$10.1} & 31.1\small{$\pm$12.3} & 56.3\small{$\pm$8.4} \\
                                           & Tail-GNN          & 64.6\small{$\pm$3.6} & 62.0\small{$\pm$5.4} & \multicolumn{1}{c|}{77.9\small{$\pm$2.4}} & 57.2\small{$\pm$1.5} & 51.8\small{$\pm$1.8} & 72.9\small{$\pm$1.1} & 55.9\small{$\pm$4.5} & 54.0\small{$\pm$5.0} & \multicolumn{1}{c|}{71.9\small{$\pm$3.2}} & 41.7\small{$\pm$1.4} & 36.8\small{$\pm$2.8} & 61.4\small{$\pm$1.1} \\ \cline{2-14} 
                                           & \textbf{\proposed}    & \textbf{73.2}\small{$\pm$5.4} & \textbf{72.1}\small{$\pm$6.1} & \multicolumn{1}{c|}{\textbf{83.6}\small{$\pm$3.5}} & \textbf{70.9}\small{$\pm$2.5} & \textbf{69.6}\small{$\pm$2.8} & \textbf{82.1}\small{$\pm$1.6} & \textbf{75.4}\small{$\pm$5.6} & \textbf{75.4}\small{$\pm$5.4} & \multicolumn{1}{c|}{\textbf{85.0}\small{$\pm$3.6}} & \textbf{70.2}\small{$\pm$4.5} & \textbf{68.8}\small{$\pm$4.7} & \textbf{81.7}\small{$\pm$3.0} \\ \hline
\multirow{12}{*}{\rotatebox{90}{CiteSeer}} & Origin            & 49.5\small{$\pm$2.1} & 43.1\small{$\pm$2.3} & \multicolumn{1}{c|}{66.7\small{$\pm$1.5}} & 48.2\small{$\pm$0.9} & 39.3\small{$\pm$0.4} & 65.7\small{$\pm$0.7} & 48.9\small{$\pm$1.4} & 45.3\small{$\pm$1.3} & \multicolumn{1}{c|}{66.2\small{$\pm$1.1}} & 42.4\small{$\pm$6.5} & 39.1\small{$\pm$7.3} & 61.1\small{$\pm$5.1} \\
                                           & Over-sampling     & 51.5\small{$\pm$3.0} & 43.7\small{$\pm$2.1} & \multicolumn{1}{c|}{68.2\small{$\pm$2.2}} & 47.8\small{$\pm$0.8} & 38.9\small{$\pm$1.9} & 65.4\small{$\pm$0.6} & 43.0\small{$\pm$3.4} & 40.3\small{$\pm$1.7} & \multicolumn{1}{c|}{61.7\small{$\pm$2.7}} & 40.1\small{$\pm$2.0} & 34.2\small{$\pm$1.5} & 59.4\small{$\pm$1.6} \\
                                           & Re-weight         & 52.1\small{$\pm$2.7} & 46.2\small{$\pm$3.2} & \multicolumn{1}{c|}{68.6\small{$\pm$2.0}} & 48.0\small{$\pm$0.4} & 39.2\small{$\pm$1.1} & 65.6\small{$\pm$0.3} & 48.4\small{$\pm$3.9} & 44.5\small{$\pm$3.9} & \multicolumn{1}{c|}{65.8\small{$\pm$2.9}} & 41.3\small{$\pm$4.5} & 35.6\small{$\pm$5.3} & 60.3\small{$\pm$3.6} \\
                                           & SMOTE             & 48.7\small{$\pm$2.5} & 40.1\small{$\pm$1.8} & \multicolumn{1}{c|}{66.1\small{$\pm$1.9}} & 47.8\small{$\pm$0.8} & 38.9\small{$\pm$1.9} & 65.4\small{$\pm$0.6} & 44.9\small{$\pm$4.4} & 41.9\small{$\pm$4.1} & \multicolumn{1}{c|}{63.2\small{$\pm$3.4}} & 40.1\small{$\pm$2.0} & 34.2\small{$\pm$1.5} & 59.4\small{$\pm$1.6} \\
                                           & Embed-SMOTE       & 47.5\small{$\pm$2.1} & 37.9\small{$\pm$1.7} & \multicolumn{1}{c|}{65.2\small{$\pm$1.6}} & 46.7\small{$\pm$3.0} & 35.7\small{$\pm$2.8} & 64.5\small{$\pm$2.3} & 43.2\small{$\pm$6.5} & 38.3\small{$\pm$5.8} & \multicolumn{1}{c|}{61.8\small{$\pm$5.2}} & 33.2\small{$\pm$6.6} & 28.3\small{$\pm$7.9} & 53.4\small{$\pm$5.9} \\
                                           & $\text{GraphSMOTE}_{T}$    & 51.2\small{$\pm$3.7} & 43.4\small{$\pm$4.2} & \multicolumn{1}{c|}{67.9\small{$\pm$2.8}} & 49.3\small{$\pm$2.0} & 40.1\small{$\pm$1.3} & 66.5\small{$\pm$1.5} & 50.3\small{$\pm$5.0} & 46.1\small{$\pm$4.5} & \multicolumn{1}{c|}{67.2\small{$\pm$3.7}} & 46.5\small{$\pm$3.7} & \textbf{41.5}\small{$\pm$4.1} & 64.4\small{$\pm$2.9} \\
                                           & $\text{GraphSMOTE}_{O}$    & 52.7\small{$\pm$2.3} & 45.3\small{$\pm$2.8} & \multicolumn{1}{c|}{69.1\small{$\pm$1.7}} & 49.5\small{$\pm$2.6} & 40.3\small{$\pm$1.8} & 66.7\small{$\pm$2.0} & 49.5\small{$\pm$3.5} & 44.5\small{$\pm$2.9} & \multicolumn{1}{c|}{66.7\small{$\pm$2.6}} & 42.3\small{$\pm$6.6} & 36.9\small{$\pm$6.6} & 61.0\small{$\pm$5.3} \\
                                           & $\text{GraphSMOTE}_{preT}$ & 44.7\small{$\pm$1.7} & 37.3\small{$\pm$2.1} & \multicolumn{1}{c|}{63.1\small{$\pm$1.3}} & 48.2\small{$\pm$3.9}                     & 39.4\small{$\pm$4.9} & 65.7\small{$\pm$3.0} & 41.8\small{$\pm$4.1} & 39.5\small{$\pm$4.1} & \multicolumn{1}{c|}{60.7\small{$\pm$3.3}} & 38.0\small{$\pm$2.6} & 33.6\small{$\pm$2.5} & 57.7\small{$\pm$2.1} \\
                                           & $\text{GraphSMOTE}_{preO}$ & 44.1\small{$\pm$2.0} & 36.6\small{$\pm$1.7} & \multicolumn{1}{c|}{62.6\small{$\pm$1.6}} & 45.7\small{$\pm$2.6} & 37.1\small{$\pm$3.1} & 63.8\small{$\pm$2.0} & 43.4\small{$\pm$6.6} & 42.9\small{$\pm$6.3} & \multicolumn{1}{c|}{62.7\small{$\pm$5.1}} & 39.2\small{$\pm$1.8} & 34.7\small{$\pm$2.4} & 58.7\small{$\pm$1.5} \\
                                           & GraphENS          & 44.2\small{$\pm$3.5} & 35.9\small{$\pm$1.0} & \multicolumn{1}{c|}{62.7\small{$\pm$2.7}} & 43.5\small{$\pm$2.6} & 33.4\small{$\pm$1.9} & 62.1\small{$\pm$2.1} & 33.0\small{$\pm$3.2} & 28.6\small{$\pm$4.4} & \multicolumn{1}{c|}{53.4\small{$\pm$2.9}} & 28.5\small{$\pm$6.7} & 23.1\small{$\pm$6.2} & 49.1\small{$\pm$6.2} \\
                                           & Tail-GNN          & 48.8\small{$\pm$1.9} & 40.4\small{$\pm$2.9} & \multicolumn{1}{c|}{66.2\small{$\pm$1.5}} & 48.2\small{$\pm$1.7} & 39.4\small{$\pm$1.2} & 65.7\small{$\pm$1.3} & 42.4\small{$\pm$6.1} & 38.9\small{$\pm$6.1} & \multicolumn{1}{c|}{61.1\small{$\pm$4.8}} & 34.2\small{$\pm$4.8} & 28.2\small{$\pm$4.1} & 54.4\small{$\pm$4.2} \\ \cline{2-14} 
                                           & \textbf{\proposed}    & \textbf{54.2}\small{$\pm$4.5} & \textbf{51.8}\small{$\pm$4.1} & \multicolumn{1}{c|}{\textbf{70.2}\small{$\pm$3.3}} & \textbf{52.7}\small{$\pm$2.1} & \textbf{48.3}\small{$\pm$3.7} & \textbf{69.1}\small{$\pm$1.5} & \textbf{52.1}\small{$\pm$3.7} & \textbf{47.2}\small{$\pm$3.6} & \multicolumn{1}{c|}{\textbf{68.6}\small{$\pm$2.7}} & \textbf{47.3}\small{$\pm$1.1} & 41.2\small{$\pm$2.1} & \textbf{65.0}\small{$\pm$0.9} \\ \hline
\end{tabular}
}
\vspace{-2ex}
\end{table*}

\begin{table}[t]
\centering
\caption{\label{tab:table4}The overall performance on \textit{manual LT} dataset (\textit{Imbalance\_ratio: 1\%}).}
\vspace{-2ex}
\resizebox{0.94\columnwidth}{!}{
\begin{tabular}{l|ccc|ccc}
\hline
\multicolumn{1}{c|}{\multirow{2}{*}{\textbf{Method}}} & \multicolumn{3}{c|}{\textbf{Cora-LT}}                              & \multicolumn{3}{c}{\textbf{CiteSeer-LT}}                           \\ \cline{2-7} 
\multicolumn{1}{c|}{}                                 & bAcc.                & Macro-F1             & G-Means              & bAcc.                & Macro-F1             & G-Means              \\ \hline \hline
Origin                                                & 66.8\small{$\pm$1.1} & 65.0\small{$\pm$1.0} & 79.5\small{$\pm$0.7} & 50.4\small{$\pm$1.4} & 45.7\small{$\pm$0.8} & 67.4\small{$\pm$1.1} \\
Over-sampling                                         & 66.6\small{$\pm$0.8} & 64.5\small{$\pm$0.6} & 79.3\small{$\pm$0.5} & 51.7\small{$\pm$1.2} & 46.7\small{$\pm$0.8} & 68.4\small{$\pm$0.9} \\
Re-weight                                             & 68.0\small{$\pm$0.7} & 66.7\small{$\pm$1.4} & 80.2\small{$\pm$0.5} & 53.0\small{$\pm$2.4} & 48.8\small{$\pm$2.3} & 69.3\small{$\pm$1.7} \\
SMOTE                                                 & 66.8\small{$\pm$0.4} & 66.4\small{$\pm$0.6} & 79.4\small{$\pm$0.2} & 51.2\small{$\pm$1.9} & 46.6\small{$\pm$1.8} & 68.0\small{$\pm$1.4} \\
Embed-SMOTE                                           & 65.2\small{$\pm$0.6} & 63.1\small{$\pm$0.3} & 78.4\small{$\pm$0.4} & 52.6\small{$\pm$1.6} & 48.0\small{$\pm$1.5} & 69.0\small{$\pm$1.2} \\
$\text{GraphSMOTE}_{T}$                                        & 67.7\small{$\pm$1.2} & 66.3\small{$\pm$1.1} & 80.0\small{$\pm$0.8} & 51.5\small{$\pm$0.7} & 46.8\small{$\pm$0.9} & 68.2\small{$\pm$0.5} \\
$\text{GraphSMOTE}_{O}$                                        & 67.1\small{$\pm$1.8} & 65.1\small{$\pm$1.9} & 79.6\small{$\pm$1.2} & 51.7\small{$\pm$0.6} & 46.8\small{$\pm$1.0} & 68.4\small{$\pm$0.5} \\
$\text{GraphSMOTE}_{preT}$                                     & 68.1\small{$\pm$1.2} & 66.6\small{$\pm$1.7} & 80.3\small{$\pm$0.8} & 47.6\small{$\pm$0.7} & 42.8\small{$\pm$0.7} & 65.3\small{$\pm$0.6} \\
$\text{GraphSMOTE}_{preO}$                                     & 67.9\small{$\pm$1.7} & 65.8\small{$\pm$2.0} & 80.1\small{$\pm$1.1} & 48.8\small{$\pm$2.0} & 44.2\small{$\pm$2.2} & 66.2\small{$\pm$1.5} \\
GraphENS                                              & 67.4\small{$\pm$1.5} & 64.5\small{$\pm$1.5} & 79.8\small{$\pm$1.0} & 52.4\small{$\pm$1.7} & 47.2\small{$\pm$1.0} & 68.9\small{$\pm$1.3} \\
Tail-GNN                                              & 67.7\small{$\pm$2.5} & 64.2\small{$\pm$2.7} & 80.0\small{$\pm$1.6} & 51.7\small{$\pm$0.4} & 46.7\small{$\pm$0.5} & 68.4\small{$\pm$0.3} \\ \hline
\textbf{\proposed}                                         & \textbf{72.2}\small{$\pm$3.1} & \textbf{72.0}\small{$\pm$2.9} & \textbf{83.0}\small{$\pm$2.0} & \textbf{56.4}\small{$\pm$2.1} & \textbf{52.5}\small{$\pm$2.2} & \textbf{71.7}\small{$\pm$1.5}  \\ \hline      
\end{tabular}
}
\vspace{-3ex}
\end{table}

\subsection{Performance Analysis}

\subparagraph{\textbf{Overall Evaluation.}} Table~\ref{tab:table3} shows the performance of node classification under various imbalanced settings where the number of imbalance classes among total classes varies at the same time. We have the observations: \textbf{1)}~\proposed~generally performs well on all datasets under various imbalance settings, especially outperforming other baselines on harsh imbalance settings such as when the number of imbalance classes is 5 and \textit{imbalance\_ratio} is 5\%. More precisely, taking Cora dataset as an example, with 7 classes in total, the available training samples for each class are 20, 20, 1, 1, 1, 1, 1. Despite its harsh condition that the last five classes are trained only once,~\proposed~notably outperforms all the oversampling-based baselines. This implies that rather than introducing oversampled synthetic samples, the balanced approach of~\proposed~that obtains relatively balanced subsets (e.g., separates the above case as 40 samples to the Head class and the remaining 5 samples to the Tail class) alleviates the class long-tailedness, and plays a significant role for alleviating the long-tail problem. \textbf{2)} We observe that when \textit{imbalance\_ratio} becomes harsh (e.g., train samples for the Tail classes reduce from 2 (i.e., 20 $\times$ 0.1) to 1 (i.e., 20 $\times$ 0.05)), most oversampling-based methods fail to generalize well. Especially, the state-of-the-art method, GraphENS, fails on such harsh settings. This is because it works by combining a target ego-network (which can be generated from any classes) with a minor ego-network (which is generated from the minority classes), and thus sampled nodes would be sampled again with high probabilities in such harsh settings, thereby failing to obtain nodes with generalized embeddings. On the other hand, GraphENS works well under the settings where both the Head class nodes and the Tail class nodes are abundant. To verify this trend, we evaluate all the above models on their proposed settings, i.e., Cora-\textit{LT} and CiteSeer-\textit{LT}. As in Table~\ref{tab:table4}, we observe that GraphENS performs competitively on their proposed setting where the Head and Tail class nodes are more abundant than our setting. However,~\proposed~still outperforms the state-of-the-art methods, which demonstrates the effectiveness of~\proposed~in various environments. \textbf{3)} Moreover, we observe that Tail-GNN, which aims to obtain high-quality embeddings for low-degree nodes, is not competitive when the \textit{imbalance\_ratio} is harsh. As a result, we can conclude that in addition to the robust modeling on the tail degree nodes, the class long-tailedness should not be overlooked. \textbf{4)} In Table~\ref{tab:table5}, we observe that~\proposed~performs well not only on the manual imbalance settings, but also on the natural imbalanced dataset (i.e., Cora-Full) especially when the number of total classes is large and the \textit{imbalance\_ratio} is harsh.

\begin{table}
\centering
\caption{\label{tab:table5}The overall performance on \textit{natural} dataset (\textit{Imbalance\_ratio: 1.1\%}) (OOM: Out Of Memory).}
\vspace{-2ex}
\resizebox{0.68\columnwidth}{!}{
\begin{tabular}{l|cccc}
\hline
\multicolumn{1}{c|}{\multirow{2}{*}{\textbf{Method}}} & \multicolumn{4}{c}{\textbf{Cora-Full}}                                                                                                          \\ \cline{2-5} 
                              & bAcc.                & Macro-F1             & G-Means              & Acc.                   \\ \hline \hline
Origin                                                & 52.0\small{$\pm$1.0} & 52.5\small{$\pm$0.8} & 71.9\small{$\pm$0.7} & 60.5\small{$\pm$0.2}  \\
Over-sampling                                         & 51.4\small{$\pm$1.0} & 52.4\small{$\pm$0.9} & 71.5\small{$\pm$0.7} & 60.9\small{$\pm$0.3}  \\
Re-weight                                             & 52.1\small{$\pm$0.9} & 52.6\small{$\pm$0.7} & 72.0\small{$\pm$0.6} & 60.7\small{$\pm$0.1}  \\
SMOTE                                                 & 52.0\small{$\pm$0.7} & 52.6\small{$\pm$0.6} & 71.9\small{$\pm$0.5} & 60.7\small{$\pm$0.1}  \\
Embed-SMOTE                                           & 52.3\small{$\pm$0.7} & \textbf{53.8}\small{$\pm$0.7} & 72.1\small{$\pm$0.5} & 62.6\small{$\pm$0.5} \\
$\text{GraphSMOTE}_{T}$                                        & 52.1\small{$\pm$0.9} & 52.4\small{$\pm$0.7} & 72.0\small{$\pm$0.6} & 60.6\small{$\pm$0.3} \\
$\text{GraphSMOTE}_{O}$                                        & 52.0\small{$\pm$0.9} & 52.4\small{$\pm$0.8} & 71.9\small{$\pm$0.6} & 60.7\small{$\pm$0.5}  \\
$\text{GraphSMOTE}_{preT}$                                     & 48.0\small{$\pm$2.1} & 48.4\small{$\pm$2.2} & 69.0\small{$\pm$1.5} & 56.8\small{$\pm$1.9}  \\
$\text{GraphSMOTE}_{preO}$                                     & 47.7\small{$\pm$1.7} & 47.7\small{$\pm$1.6} & 68.8\small{$\pm$1.3} & 56.3\small{$\pm$1.5}  \\
GraphENS                                              & 49.6\small{$\pm$0.6} & 51.5\small{$\pm$0.5} & 70.2\small{$\pm$0.4} & \textbf{62.5}\small{$\pm$0.3}  \\
Tail-GNN                                              & OOM                  & OOM                  & OOM                  & OOM                   \\ \hline
\textbf{\proposed}                                        & \textbf{54.2}\small{$\pm$0.7} & 53.0\small{$\pm$0.4} &\textbf{73.4}\small{$\pm$0.5} & 60.9\small{$\pm$0.5}   \\ \hline
\end{tabular}
}
\vspace{-5.5ex}
\end{table}

\begin{figure*}[t]
\centering
    \includegraphics[width=1.4\columnwidth]{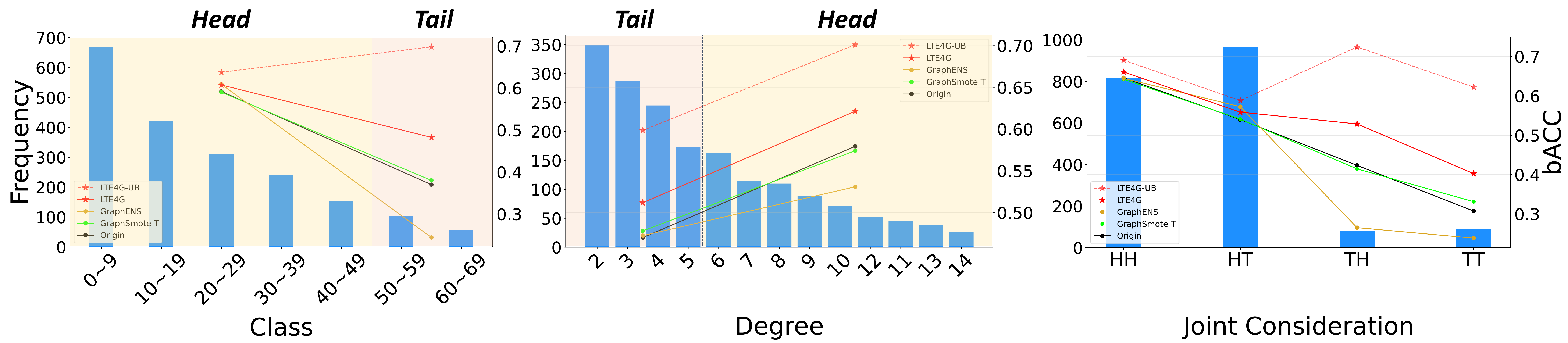}
    \vspace{-2ex}
    \caption{Performance on each class, degree and joint consideration measured in Cora-Full dataset.~\proposed\textsf{-UB} denotes a variant of~\proposed~when all the test nodes are assumed to be assigned to correct class-wise students.}
    \label{fig:figure4}
    \vspace{-2ex}
\end{figure*}

\subsection{Ablation Studies}

\noindent{\textbf{Benefit of Each Component in~\proposed. }} Table~\ref{tab:table6} shows the node classification performance when considering (a) class long-tailedness only, (b) degree long-tailedness only, (c) both class and degree long-tailedness, (d) with knowledge distillation, (e) with tail-to-head curriculum learning and (f) with head-to-tail curriculum learning (\proposed). We have the following observations:
\textbf{1)} Considering only the class-longtailedness performs better than considering only the degree-longtailedness ((a)$>$(b)).
\textbf{2)} Considering either one of the class-longtailedness or the degree-longtailedness is not enough, and even jointly considering them performs worse than considering only the class-longtailedness ((a)$>$(c)). 
However, as the knowledge distillation is introduced, the performance increases ((d)$>$(c)). This implies 
that the knowledge in each expert is not fully utilized unless knowledge is properly distilled to the students.
\textbf{3)} While training each student, i.e, Head class student and Tail class student, degree wise head-to-tail learning improves the overall performance ((f)$>$(e)).
\textbf{4)}~\proposed~outperforms all other variants demonstrating the superiority of our knowledge distillation framework.
\textbf{5)} In Figure~\ref{fig:figure4}, we plot bAcc for each of the balanced subsets and we found that our expert-based model,~\proposed~outperforms other imbalanced methods (e.g., GraphSMOTE, GraphENS) in terms of class perspective (Figure~\ref{fig:figure4}(a)), degree perspective (Figure~\ref{fig:figure4}(b)), and their joint consideration (Figure~\ref{fig:figure4}(c)).


\begin{table}[t]
\centering
\caption{\label{tab:table6} Ablations on each component of~\proposed. $C$, $D$, $KD$, $T2H$ and $H2T$ denote consideration of class, degree, knowledge distillation (i.e., student) Tail-to-Head and Head-to-Tail curriculum learning, respectively. The last row:~\proposed.}
\vspace{-2ex}
\resizebox{\columnwidth}{!}{%
\begin{tabular}{cccccccccccc}
\hline
\multicolumn{6}{c|}{\textbf{Components}}                                                                                                                                                & \multicolumn{3}{c|}{\textbf{CiteSeer-10\% (Imb. Class 5)}}                                                                & \multicolumn{3}{c}{\textbf{CiteSeer-5\% (Imb. Class 5)}}                                            \\ \hline
\multicolumn{1}{c|}{\#}          & \textit{C}              & \textit{D}              & \textit{KD}             & \textit{T2H}            & \multicolumn{1}{c|}{\textit{H2T}}            & bAcc.                         & Macro-F1                      & \multicolumn{1}{c|}{G-Means}                       & bAcc.                         & Macro-F1                      & G-Means                       \\ \hline \hline
\multicolumn{1}{c|}{(a)}          & \color{green}\checkmark &                         &                         &                         & \multicolumn{1}{c|}{}                        & 51.5\small{$\pm$4.2}          & 47.1\small{$\pm$4.0}          & \multicolumn{1}{c|}{68.2\small{$\pm$3.1}}          & 43.5\small{$\pm$0.5}          & 38.5\small{$\pm$1.8}          & 62.1\small{$\pm$0.4}          \\
\multicolumn{1}{c|}{(b)}          &                         & \color{green}\checkmark &                         &                         & \multicolumn{1}{c|}{}                        & 39.6\small{$\pm$4.6}          & 34.7\small{$\pm$5.8}          & \multicolumn{1}{c|}{58.9\small{$\pm$3.8}}          & 29.8\small{$\pm$2.5}          & 24.1\small{$\pm$2.4}          & 50.6\small{$\pm$2.3}          \\
\multicolumn{1}{c|}{(c)}          & \color{green}\checkmark & \color{green}\checkmark &                         &                         & \multicolumn{1}{c|}{}                        & 45.6\small{$\pm$2.3}          & 41.1\small{$\pm$3.2}          & \multicolumn{1}{c|}{63.7\small{$\pm$1.8}}          & 37.6\small{$\pm$5.7}          & 32.9\small{$\pm$5.8}          & 57.2\small{$\pm$4.8}          \\
\multicolumn{1}{c|}{(d)}          & \color{green}\checkmark & \color{green}\checkmark & \color{green}\checkmark &                         & \multicolumn{1}{c|}{}                        & 50.7\small{$\pm$3.3}          & 45.5\small{$\pm$2.8}          & \multicolumn{1}{c|}{67.6\small{$\pm$2.5}}          & 44.9\small{$\pm$3.6}          & 39.4\small{$\pm$1.5}          & 63.2\small{$\pm$2.8}          \\
\multicolumn{1}{c|}{(e)}          & \color{green}\checkmark & \color{green}\checkmark & \color{green}\checkmark & \color{green}\checkmark & \multicolumn{1}{c|}{}                        & 50.5\small{$\pm$2.8}          & 45.9\small{$\pm$2.0}          & \multicolumn{1}{c|}{67.4\small{$\pm$2.1}}          & 46.5\small{$\pm$3.1}          & 41.9\small{$\pm$3.7}          & 64.4\small{$\pm$2.3}          \\ \hline
\multicolumn{1}{c|}{\textbf{(f)}} & \color{green}\checkmark & \color{green}\checkmark & \color{green}\checkmark &                         & \multicolumn{1}{c|}{\color{green}\checkmark} & \textbf{52.1\small{$\pm$3.7}} & \textbf{47.2\small{$\pm$3.6}} & \multicolumn{1}{c|}{\textbf{68.6\small{$\pm$2.7}}} & \textbf{47.3\small{$\pm$1.1}} & \textbf{41.2\small{$\pm$2.1}} & \textbf{65.0\small{$\pm$0.9}} \\ \hline
\multicolumn{1}{l}{}             & \multicolumn{1}{l}{}    & \multicolumn{1}{l}{}    & \multicolumn{1}{l}{}    & \multicolumn{1}{l}{}    & \multicolumn{1}{l}{}                         & \multicolumn{1}{l}{}          & \multicolumn{1}{l}{}          & \multicolumn{1}{l}{}                               & \multicolumn{1}{l}{}          & \multicolumn{1}{l}{}          & \multicolumn{1}{l}{}         
\end{tabular}
}
\vspace{-3ex}
\end{table}

\begin{figure}[t]
\includegraphics[width=0.7\columnwidth]{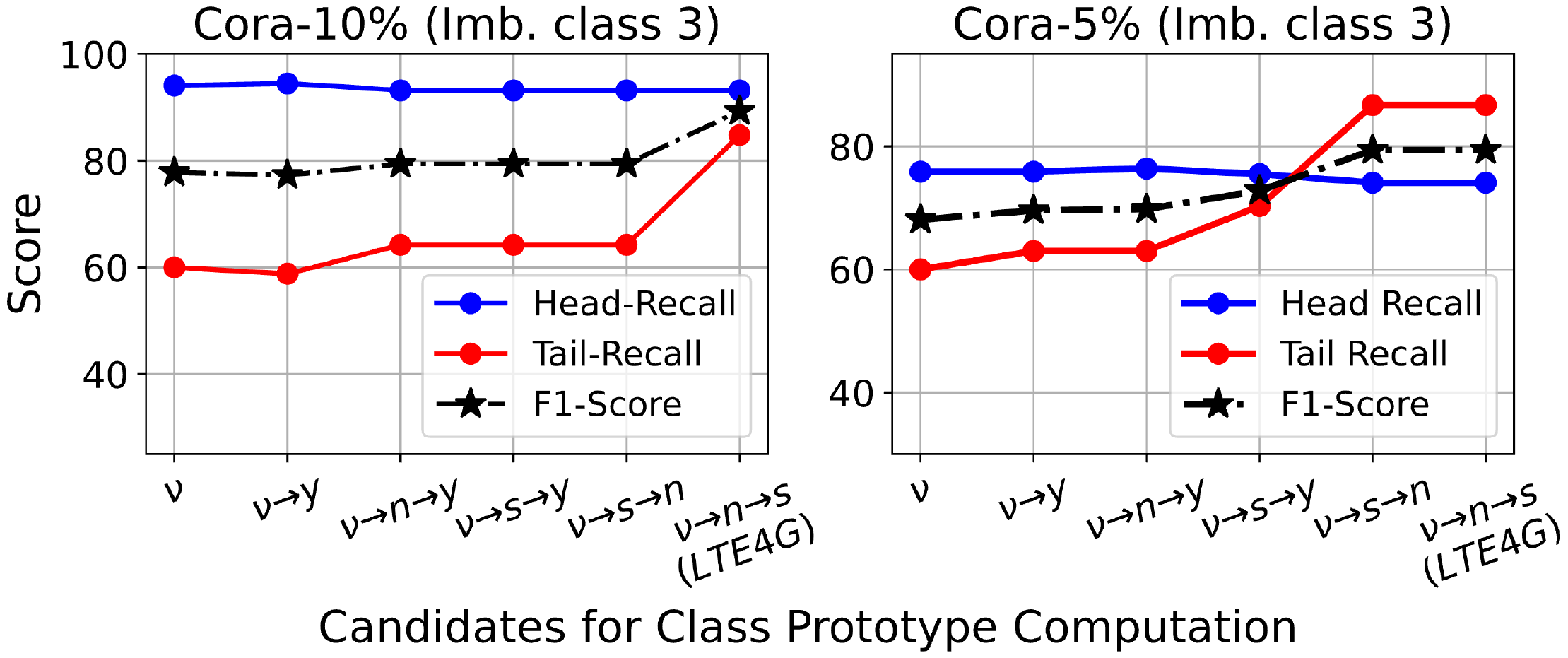}
\vspace{-3ex}
\caption{Performance of student assignments across ablations of candidates for class prototype computation. $v$, $n$, $s$, $y$ denote node ($\mathcal{V}_\text{train}$), neighboring nodes ($\mathcal{N}_{\text{train}}$), top-$k$ similar nodes ($\mathcal{S}_{\text{train}}$), and nodes to supplement candidates that have high probabilities belonging to the class, respectively. "$\rightarrow$" denotes the order of candidate selection process.
}
\label{fig:figure5}
\vspace{-4.25ex}
\end{figure}

\smallskip
\noindent{\textbf{Ablations on Class Prototype-based Inference.}} 
Figure~\ref{fig:figure5} shows ablations on candidates for class  prototype computation described in Sec.~\ref{sec:3.6}. 
Here, $y$-axis shows the performance of the student assignment. Our aim is to find the variant that provides the best F1-score since we need to assign test nodes to both Head- and Tail-class students relatively well. We observe that \textbf{1)} ``$v\rightarrow y$'' outperforms ``$v$'' implying that incorporating nodes with high probabilities belonging to the same class as the target node as candidates is helpful for obtaining high-quality class prototypes rather than solely considering the target node.
\textbf{2)} "$v\rightarrow n(s)\rightarrow y$" outperforms "$v\rightarrow y$'', which further reveals that considering neighboring or similar nodes in prior is helpful. \textbf{3)} "$v\rightarrow n\rightarrow s$" outperforms "$v\rightarrow s\rightarrow n$" implying that \textit{order matters}. To be specific, when incorporating candidate nodes, considering the neighboring nodes in prior is more important than considering similar nodes as prior.

\begin{table}[h]
\caption{\label{tab:table7} Ablations on balancing node distribution
($\color{green}\checkmark$: balanced split, $\color{red}\xmark$: random split). The last row:~\proposed.
}
\vspace{-2ex}
\resizebox{\columnwidth}{!}{%
\begin{tabular}{cc|ccc|ccc}
\hline 
\multicolumn{2}{c|}{\textbf{Balanced Split}} & \multicolumn{3}{c|}{\textbf{Cora-5\% (Imb. Class 3)}} & \multicolumn{3}{c}{\textbf{CiteSeer-5\% (Imb. Class 3)}} \\ \hline 
Class          & Degree         & bAcc.  & Macro-F1  & G-Means & bAcc.  & Macro-F1  & G-Means  \\ \hline \hline
\color{red}\xmark               & \color{red}\xmark              & 51.9\small{$\pm$3.5}      & 44.4\small{$\pm$4.8}         & 69.1\small{$\pm$2.5}       & 38.1\small{$\pm$1.9}      & 26.6\small{$\pm$0.5}         & 57.7\small{$\pm$1.5}        \\
\color{red}\xmark              &       \color{green}\checkmark       &  47.4\small{$\pm$2.0}     & 36.7\small{$\pm$2.2}         & 65.8\small{$\pm$1.5}       & 38.8\small{$\pm$1.7}      & 27.0\small{$\pm$1.0}         & 58.3\small{$\pm$1.4}        \\
    \color{green}\checkmark           &\color{red}\xmark             & 69.4\small{$\pm$2.6}      & 68.2\small{$\pm$2.8}        & 81.2\small{$\pm$1.7}       & 52.2\small{$\pm$1.6}      & 48.2\small{$\pm$3.1}         & 68.7\small{$\pm$1.2}       \\
 \hline
\color{green}\checkmark              &    \color{green}\checkmark           & \textbf{70.9}\small{$\pm$2.5} & \textbf{69.6}\small{$\pm$2.8} & \textbf{82.1}\small{$\pm$1.6}
& \textbf{52.7}\small{$\pm$2.1} & \textbf{48.3}\small{$\pm$3.7} &\textbf{69.1}\small{$\pm$1.5}        \\ \hline
\end{tabular}%
}
\label{tab:balance}
\vspace{-2ex}
\end{table}

\smallskip
\noindent{\textbf{Ablations on Balancing Node Distribution.}}
To verify the benefit of balancing the node distribution considering both the class and the degree distributions as described in Sec.~\ref{sec:3.2}, we evaluate the performance of~\proposed~when either one of them (or both) is not balancedly considered (Table~\ref{tab:balance}). For instance, in third row of Table ~\ref{tab:table7}~, ($\color{green}\checkmark$,$\color{red}\xmark$) means the nodes are split into two subsets considering the class distribution, and then each subset is further \textit{randomly} split into two subsets rather than using the degree criterion (i.e., node degree above 5 or not) that splits the Head class into HH or HT. We observe that 
\textbf{1)} randomly splitting nodes regardless of the class and the degree distributions results in the experts not being able to fully obtain their specialized knowledge from the subsets (i.e., random splitting does not guarantee that each subset is class- and degree-wise balanced). For this reason, even if knowledge distillation is performed, the knowledge distilled from experts to the class-wise students is not informative.
\textbf{2)} Among class and degree-balanced split, class-balanced split is more effective than degree-balanced split which turns out to be negatively affect overall performance when class long-tailedness remains. 
\textbf{3)} All in all, the beauty of alleviating long-tailedness comes in where \textit{the both class and degree long-tailedness is jointly considered}.




\subsection{Sensitivity \& Complexity Analysis}

\begin{figure}[t]
\includegraphics[width=0.83\columnwidth]{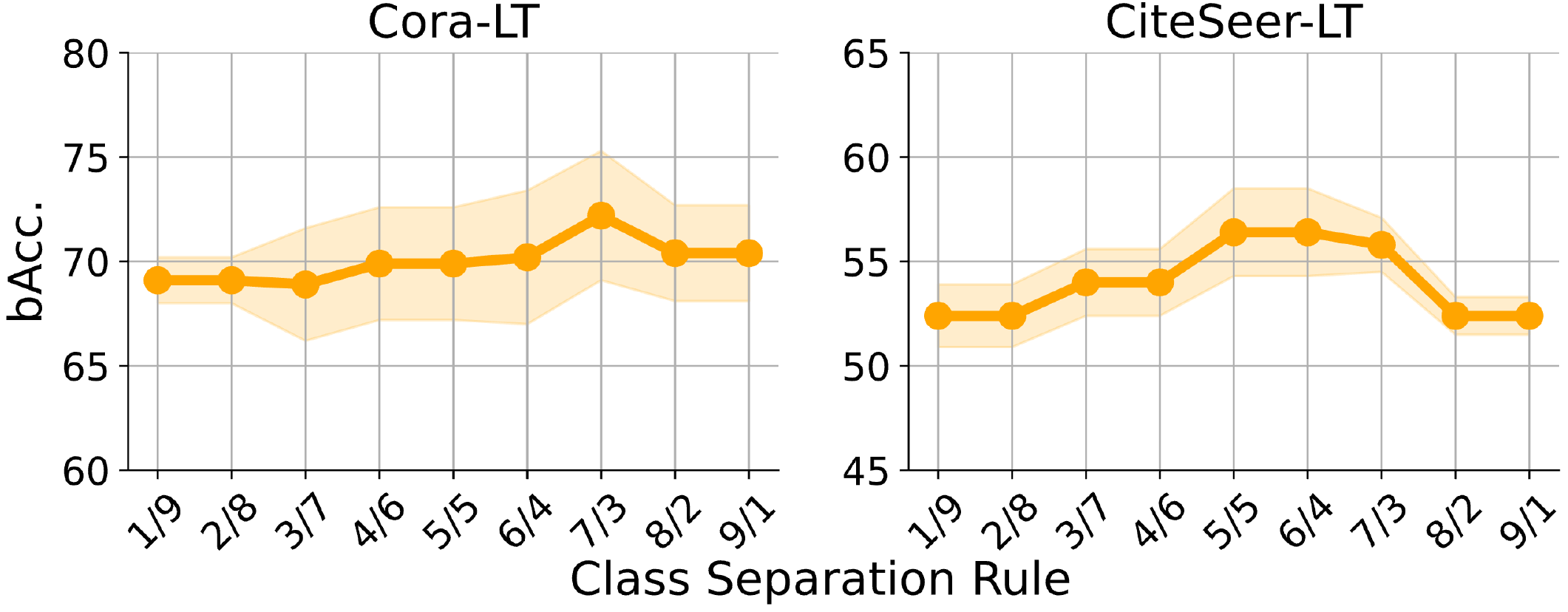}
\vspace{-2ex}
\caption{\label{fig:figure6}Sensitivity analysis on class separation.}
\vspace{-2ex}
\end{figure}

\noindent{\textbf{Hyperparameter Analysis.}} 
\looseness=-1
Figure~\ref{fig:figure6} shows the sensitivity analysis on an important hyperparameter of~\proposed, i.e., the separation rule for splitting the classes in a balanced manner (i.e., portions of head classes and tail classes). We observe that regarding the separation rule, best-performing separation usually differs since the cardinality of each dataset varies. Nonetheless, regarding the head class as over top-$60\%$, and remaining as tail class works well in general.

\begin{figure}[t]
\includegraphics[width=0.83\columnwidth]{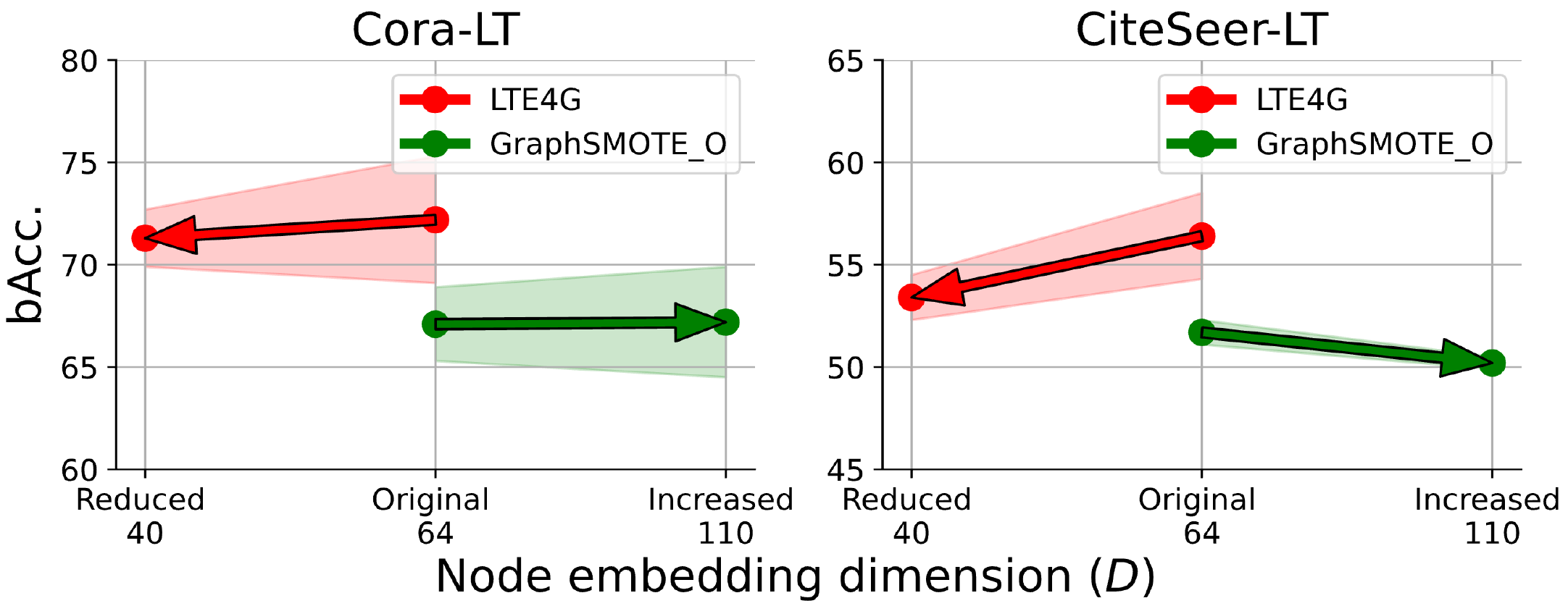}
\vspace{-2ex}
\caption{Complexity analysis on the number of parameters.}
\label{fig:figure7}
\vspace{-3ex}
\end{figure}

\smallskip
\noindent{\textbf{Complexity Analysis.}} Since our framework contains more GNN-based classifiers (i.e., 1-layer GNN followed by 1-layer MLP) due to the experts (e.g., HH, HT, TH, TT experts), it is reasonable to conduct complexity analysis in terms of the total number of parameters used for training. Among baselines, we choose GraphSMOTE as a compared model and conduct a complexity analysis.
To this end, we either i) increase the size of the node embedding dimension (i.e., $D$) of GraphSMOTE until the total number of parameters of GraphSMOTE equals that of~\proposed, or ii) decrease $D$ of~\proposed~until the total number of parameters of~\proposed~equals that of GraphSMOTE.
In Figure~\ref{fig:figure7}, we observe that even though $D$ of~\proposed~is much smaller than $D$ of GraphSMOTE,~\proposed~still outperforms both the original GraphSMOTE and its dimension-increased version, which demonstrates the superiority of our framework. 
This also implies that rather than simply increasing the number of parameters, it is important to \textit{assign parameters in the right place where they are needed}, e.g., experts considering both the class and degree long-tailedness.

\section{Related Work}
\noindent{\textbf{Imbalanced Learning. }} To handle the class-longtailedness problem where the number of samples in few classes (i.e., Head classes) dominates other classes (i.e., Tail classes), there exist classical approaches that can be categorized into three categories, e.g., data-level, algorithm-level and hybrid. First, in the data-level approach, samples that belong to minority classes are over sampled, whereas those that belong to majority classes are undersampled. 
Specifically, SMOTE~\cite{smote} generates new samples by interpolating minority samples with their nearest neighbors. 
In the algorithm-level approach, cost-sensitive methods~\cite{cost-sensitive-a, cost-sensitive-b} introduce a cost matrix that penalizes mis-classified examples (e.g., false positives and false negatives). Hybrid methods combine advantages of the data-level and algorithm-level approaches in various ways. 
Specifically, SMOTE-Boost \cite{smoteboost} combines SMOTE with boosting techniques, and there also exist methods that combine oversampling and cost sensitive learning \cite{oversample_cost}.
Recently, in computer vision domain, multi-experts based learning to handle long-tailed problem have been introduced. BbN \cite{bbn} uses two networks each of which involves normal and reversed sampling, respectively, and incorporates cumulative learning that adjusts bilateral training. To utilize more diverse classifiers, LFME \cite{lfme} and RIDE \cite{ride} propose multi-experts in terms of knowledge distillation with curriculum instance selection and distribution-aware diversity loss, respectively. \\


\smallskip
\noindent{\textbf{Imbalanced Learning on Graphs - Class perspective.}} 
\looseness=-1
Recently, few researches have been devoted to alleviating long-tail class distribution on graphs~\cite{graphsmote, imgagn, drgcn}. 
More precisely, DRGCN \cite{drgcn} and ImGAGN \cite{imgagn} are adversarial learning-based methods that generate nodes that belong to minority classes to balance the class distribution. GraphSMOTE \cite{graphsmote} adopts SMOTE on graphs by interpolating embeddings of nodes that belong to minority classes, and generates edges between the generated nodes by using an edge generator, which is pre-trained on the original graph. A recently proposed method called GraphENS \cite{graphens} uses augmentation of minor node's ego network (i.e., one-hop neighbors) with target node's ego network with feature saliency information to prevent generation of noisy samples. However, both GraphSMOTE and GraphENS rely on synthetic nodes that are generated rather than using the original graph, which might bring bias and unnecessary information into the system. What is worse is that when the number of nodes that belong to the minority class is small, e.g., a single node in an extreme case, oversampling fails as interpolating with few nodes would not generate diverse nodes.
Furthermore, recent methods tackling the long-tail class distribution of graphs overlook the degree-biased issue. \\


\smallskip
\looseness=-1
\noindent{\textbf{Imbalanced Learning on Graphs - Degree perspective.}} Addressing long-tail problem of the node degree on graphs have recently garnered attention~\cite{demonet, sl-dsgcn, metatail2vec, tailgnn}. More precisely, meta-tail2vec \cite{metatail2vec} proposes a two-stage framework that produces initial embeddings for all nodes, and further refines tail-degree nodes using the embeddings learned for head-degree nodes. 
Most recently, Tail-GNN \cite{tailgnn} aims to transfer meaningful information from high-degree nodes to low-degree nodes via neighborhood translation that captures the relational tie between a node and its neighbors. 
Despite the effectiveness of the aforementioned methods developed to address the long-tailed degree distribution of graphs, these methods pay little attention to class long-tailedness which is crucial especially when the downstream task at hand is node classification.

\section{Conclusion}
In this paper, we propose a novel GNN-based method for node classification that jointly considers the class-longtailedness and the node degree-longtailedness.
Through our empirical analysis, we observed that the fact that a node belongs to a Head class does not always imply a high classification accuracy, due to the imbalanced node degree distribution.
To this end,~\proposed~first splits the nodes in a graph into four balanced subsets considering both the class and degree distributions, and train an expert on each of the balanced subsets. Then, the knowledge distillation technique equipped with degree-wise head-to-tail student learning is employed. For inference, we devise a class prototype-based inference that utilizes not only the labeled nodes in the training set, but also their neighboring nodes along with similar nodes.
Through extensive experiments on various manual and natural imbalanced settings, we empirically demonstrated the superiority of our framework.

\smallskip
\looseness=-1
\noindent\textbf{Acknowledgement}: This work was supported by Institute of Information \& communications Technology Planning \& Evaluation (IITP) grant funded by the Korea government(MSIT) (No.2022-0-00077, 100\%)



\bibliographystyle{ACM-Reference-Format}
\balance
\bibliography{LTE4G}

\end{document}